\pdfoutput=1
\documentclass[11pt]{article}

\usepackage[final]{acl}

\usepackage{times}
\usepackage{latexsym}

\usepackage[T1]{fontenc}
\usepackage[utf8]{inputenc}

\usepackage[disable]{todonotes}

\def\fms#1{\todo[color=white]{\textbf{Fabian} #1}}
\def\lihu#1{\todo[color=yellow]{\textbf{Lihu} #1}}
\usepackage{microtype}

\usepackage{threeparttable}
\usepackage{multirow}
\usepackage{booktabs} 
\usepackage{amssymb}
\usepackage{amsmath}
\usepackage{tabu}
\usepackage{hyperref}
\usepackage{colortbl}
\usepackage{graphicx}
\usepackage{bbm}
\usepackage{makecell}
\newcommand{\ignore}[1]{}
\title{Imputing Out-of-Vocabulary Embeddings with LOVE Makes Language Models Robust with Little Cost}

\author{
Lihu Chen\textsuperscript{\rm 1},
Gaël Varoquaux\textsuperscript{\rm 2}, 
Fabian M. Suchanek\textsuperscript{\rm 1} \\
\textsuperscript{\rm 1} LTCI \& Télécom Paris \& Institut Polytechnique de Paris, France \\
\textsuperscript{\rm 2} Soda, Inria Saclay \& CEA \& Université Paris-Saclay, France \\
\texttt{\{lihu.chen, fabian.suchanek\}@telecom-paris.fr}\\ 
\texttt{\{gael.varoquaux\}@inria.fr}
}

\begin{document}
\maketitle
\begin{abstract}
State-of-the-art NLP systems represent inputs with word embeddings, but these are brittle when faced with Out-of-Vocabulary (OOV) words.
To address this issue, we follow the principle of mimick-like models to generate vectors for unseen words, by learning the behavior of pre-trained embeddings using only the surface form of words.
We present a simple contrastive learning framework, LOVE, which extends the word representation of an existing pre-trained language model (such as BERT), and makes it robust to OOV with few additional parameters.
Extensive evaluations demonstrate that our lightweight model achieves similar or even better performances than prior competitors, both on original datasets and on corrupted variants. 
Moreover, it can be used in a plug-and-play fashion with FastText and BERT, where it significantly improves their robustness.
%significantly improves the robustness of pre-trained embeddings, e.g.  to out-of-vocabulary (OOV) words.
\end{abstract}

\section{Introduction}
\begin{figure}[t]
	\centering
	\includegraphics[width=0.5\textwidth]{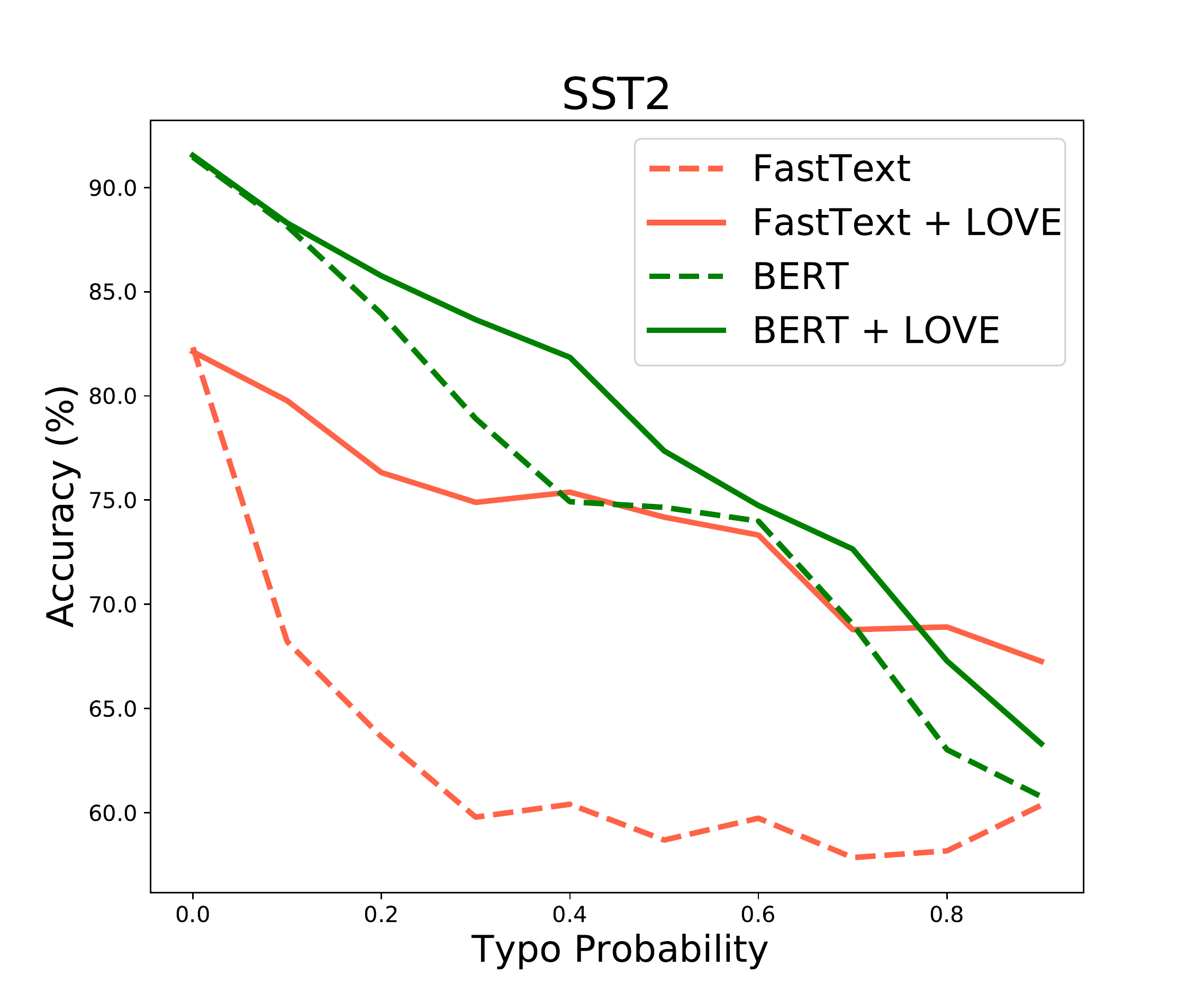}
	\caption{Performances of existing word embeddings as we gradually add typos to the datasets. Using our model, LOVE, to produce vectors for OOV words makes the models more robust.}
	\label{fig:typo_impact}
\end{figure}

Word embeddings represent words as vectors~\cite{mikolov2013efficient, mikolov2013distributed, pennington2014glove}. They have been instrumental in neural network approaches that brought impressive performance gains to many natural language processing (NLP) tasks. These approaches use a fixed-size vocabulary. Thus they can deal only with words that have been seen during training. While this works well on many benchmark datasets, real-word corpora are typically much noisier and contain Out-of-Vocabulary (OOV) words, i.e., rare words, domain-specific words, slang words, and words with typos, which have not been seen during training.
Model performance deteriorates a lot with unseen words, and  
minor character perturbations can flip the prediction of a model~\cite{liang2018deep, belinkov2018synthetic, sun2020adv,jin2020bert}.
Simple experiments (Figure~\ref{fig:typo_impact}) show that the addition of typos to datasets degrades the performance for text classification models considerably.

To alleviate this problem, pioneering work pre-trained word embeddings with morphological features (sub-word tokens) on large-scale datasets~\cite{wieting2016charagram,bojanowski2017enriching,heinzerling2017bpemb, zhang2019biowordvec}. One of the most prominent works in this direction is FastText~\cite{bojanowski2017enriching}, which incorporates character n-grams into the skip-gram model.
With FastText, vectors of unseen words can be imputed by summing up the n-gram vectors. 
However, these subword-level models come with great costs: the requirements of pre-training from scratch and high memory footprint.  Hence, several simpler approaches 
%without pre-training are developed.
have been developed, e.g., MIMICK~\cite{pinter2017mimicking}, BoS~\cite{zhao2018generalizing} and KVQ-FH~\cite{sasaki2019subword}. 
These use only the surface form of words to generate vectors for unseen words, through learning from pre-trained embeddings. 

Although MIMICK-like models can efficiently reduce the parameters of pre-trained representations and alleviate the OOV problem, two main challenges remain.
First, the models remain bound in the trade-off between complexity and performance:
%First, how to develop a small and robust enough model to cope with unseen words? 
The original MIMICK is lightweight but does not produce high-quality word vectors consistently.
BoS and KVQ-FH obtain better word representations but need more parameters.
Second, these models cannot be used with existing pre-trained language models such as BERT. It is these models, however, to which we owe so much progress in the domain~\cite{peters-etal-2018-deep,devlin2019bert,yang2019xlnet,liu2020roberta}. 
To date, %However, these advanced models such as BERT 
these high-performant models are still fragile when dealing with rare words~\cite{schick2020rare}, misspellings~\cite{sun2020adv} and domain-specific words~\cite{el2020characterbert}.

\begin{figure}[t]
	\centering
	\includegraphics[width=0.45\textwidth]{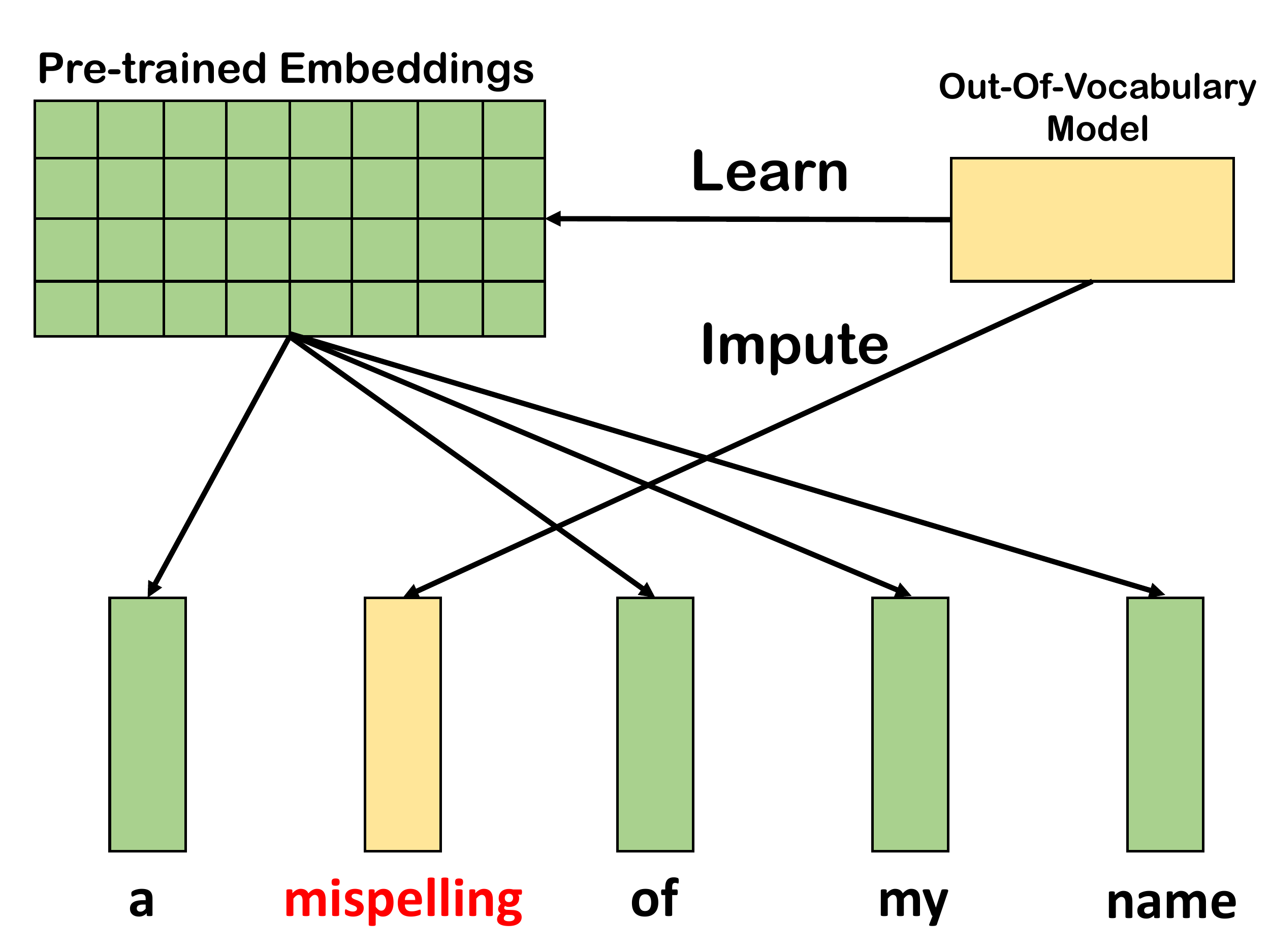}
	\caption{Our lightweight OOV model, LOVE, learns the behavior of pre-trained embeddings (e.g., FastText and BERT), and is then able to impute vectors for unseen words. LOVE can enhance the robustness of existing word representations in a plug-and-play fashion.}
	\label{fig:mimic_model}
\end{figure}

% Fabian: Put main achievements first
We address these two challenges head-on: we design a new contrastive learning framework to learn the behavior of pre-trained embeddings, dubbed LOVE, \textbf{L}earning \textbf{O}ut-of-\textbf{V}ocabulary \textbf{E}mbeddings.
Our model builds upon a memory-saving mixed input of character and subwords instead of n-gram characters. It encodes this input by a lightweight Positional Attention Module.
During training, LOVE uses novel types of data augmentation and hard negative generation. 
The model is then able to produce high-quality word representations that are robust to character perturbations,
while consuming only a fraction of the cost of existing models.
For instance, LOVE with 6.5M parameters can obtain similar representations as the original FastText model with more than 900M parameters.
What is more, our model can be used in a plug-and-play fashion to robustify existing language models. We find that using LOVE to produce vectors for unseen words improves the performance of FastText and BERT by around 1.4-6.8 percentage points on noisy text -- without hampering their original capabilities (As shown in Figure~\ref{fig:mimic_model}). 

In the following, Section~\ref{sec:related-work} discusses related work, Section~\ref{sec:background} introduces preliminaries, Section~\ref{sec:approach} presents our approach, Section~\ref{sec:experiments} shows our experiments, and Section~\ref{sec:conclusion} concludes. The appendix contains additional experiments and analyses. Our code and data is available at \url{https://github.com/tigerchen52/LOVE}

\section{Related Work}\label{sec:related-work}
\subsection{Character-level Embeddings}
To address OOV problems, some approaches inject character-level features into word embeddings during the pre-training~\cite{wieting2016charagram,cao2016joint,bojanowski2017enriching,heinzerling2017bpemb,kim2018learning, li2018subword, ustun2018characters, piktus2019misspelling,zhu2019systematic, zhang2019biowordvec, hu2019few}.
One drawback of these methods is that they need to pre-train on a large-scale corpus from scratch. 
Therefore,  simpler models have been developed, which directly mimic the well-trained word embeddings to impute vectors for OOV words. 
Some of these methods use only the surface form of words to generate embeddings for unseen words~\cite{pinter2017mimicking,zhao2018generalizing,sasaki2019subword,fukuda2020robust, jinman2020pbos},
while others use both surface and contextual information to create OOV vectors~\cite{schick2019attentive, schick2019learning}.
In both cases, the models need an excessive number of parameters. 
FastText, e.g., uses \textasciitilde2 million n-gram characters to impute vectors for unseen words.

%We follow the core idea of mimic-like models but explore how to achieve a perfect trade-off between simplicity and performance.

\subsection{Pre-trained Language Models}
Currently, the state-of-the-art word representations are pre-trained language models, such as ELMo~\cite{peters-etal-2018-deep}, BERT~\cite{devlin2019bert} and XLnet~\cite{yang2019xlnet}, which adopt subwords to avoid OOV problems.
However, BERT is brittle when faced with rare words~\cite{schick2020rare} and misspellings~\cite{sun2020adv}.
To make BERT more robust, CharacterBERT~\cite{el2020characterbert} and CharBERT~\cite{ma2020charbert} infuse character-level features into BERT and pre-train the variant from scratch. 
This method can significantly improve the performance and robustness of BERT, but requires pre-training an adapted transformer on a large amount of data.
Another work on combating spelling mistakes recommends placing a word corrector before downstream models~\cite{pruthi2019combating}, which is effective and reusable.
The main weakness of this method is that an error generated by the word corrector propagates to downstream tasks.
For example, converting \textit{``aleph''} to \textit{``alpha''} may break the meaning of a mathematical statement.
%\gv{We can also do ``bisector'' (which is not in mobile phone's vocabulary) and ``bisected''}
And indeed, using the word corrector %all backoff strategies 
consistently leads to a drop (0.5-2.0 percentage points) in BERT's performance on the SST dataset~\cite{socher2013recursive}.

\subsection{Contrastive Learning}
The origin of contrastive learning can be traced back to the work by~\citet{becker1992self} and~\citet{bromley1993signature}.
This method has achieved outstanding success in self-supervised representation learning for images~\cite{oord2018representation,hjelm2018learning,he2020momentum,chen2020simple,grill2020bootstrap}. The contrastive learning framework learns representations from unlabeled data by pulling positive pairs together and pushing negative pairs apart. 
For training, the positive pairs are often obtained by taking two randomly augmented versions of the same sample and treating the other augmented examples within a mini-batch as negative examples~\cite{chen2017sampling, chen2020simple}. The most widely used loss is the infoNCE loss (or contrastive loss)~\cite{hjelm2018learning,logeswaran2018an, chen2020simple,he2020momentum}. 
Although many approaches adopt contrastive learning to represent sentences~\cite{giorgi2020declutr,wu2020clear,gao2021simcse}, it has so far not been applied
%rare attention are paid to the area of 
to word representations.  

\begin{table}[ht] 
	\centering 
	\small
	\setlength{\tabcolsep}{2.2mm}{
	\begin{threeparttable} 
		\begin{tabular}{cccc}  
			\toprule  
			%&$\zeta(\cdot)$&$\phi(\cdot)$&$\mathcal{L}(\cdot)$&$\psi (\cdot)$\cr
			&Input&Encoder&Loss\cr
			\midrule
			%fastText \shortcite{bojanowski2017enriching} & \makecell[c]{ n-gram subword \\ \texttt{ \{spe,pel,ell\} }}&SUM &$\mathcal{L}_{\mathrm{NCE}}$ \cr
			\makecell[c]{MIMICK\\\shortcite{pinter2017mimicking}}&\makecell[c]{ character sequence \\ \texttt{ \{s,p,e,l,l\} }}&RNNs&$\mathcal{L}_{\mathrm{dis}}$\\[1em]
			\makecell[c]{BoS\\\shortcite{zhao2018generalizing}} &\makecell[c]{ n-gram subword \\ \texttt{ \{spe,pel,ell\} }}&SUM &$\mathcal{L}_{\mathrm{dis}}$  \\[1em]
			\makecell[c]{KVQ-FH\\ \shortcite{sasaki2019subword}} &\makecell[c]{ adapted n-gram subword \\ \texttt{ \{spe,pel,ell\} }}&Attention &$\mathcal{L}_{\mathrm{dis}}$ \cr
			\bottomrule
		\end{tabular}
		\caption{Details of different mimick-like models, with the word \texttt{spell} as an example.}	\label{tab:mimic-like table}
	\end{threeparttable}
	}
	%\end{minipage}%
\end{table}

\section{Preliminaries}\label{sec:background}
\subsection{Mimick-like Model} \label{sec:mimic_like_model}
Given pre-trained word embeddings, and given an OOV word, the core idea of MIMICK~\cite{pinter2017mimicking} is to impute an embedding for the OOV word using the surface form of the word, so as to mimic the behavior of the known embeddings. % and then impute vectors for unseen words. 
BoS~\cite{zhao2018generalizing}, KVQ-FH~\cite{sasaki2019subword}, Robust Backed-off Estimation~\cite{fukuda2020robust}, and PBoS~\cite{jinman2020pbos} work similarly, and we refer to them as mimick-like models.

Formally, we have a fixed-size vocabulary set $\mathcal{V}$, with 
%that defines which words are involved in this task, 
%and the corresponding 
an embedding matrix $\mathbf{W}\in \mathbb{R}^{|\mathcal{V}| \times m}$, 
in which each row is a word vector $\mathbf{u}_{w} \in \mathbb{R}^{m}$ for the word $w$.
A mimick-like model aims to impute a vector $\mathbf{v}_{w}$ for an arbitrary word $w\not\in \mathcal{V}$.
The training objective of mimick-like models is to minimize the expected distance between $\mathbf{u}_{w} $ and $\mathbf{v}_{w}$ pairs:
\begin{equation}\label{eq:loss_dis}
\mathcal{L}_{\mathrm{dis}} = \frac{1}{|\mathcal{V}|} \sum_{w \in \mathcal{V} } \psi (\mathbf{u}_{w},\mathbf{v}_{w})
\end{equation}
\noindent Here, $\psi (\cdot)$ is a distance function, e.g., the Euclidean distance $\psi  = \left \| \mathbf{u}_{w}-\mathbf{v}_{w} \right \| ^{2}_{2}$ or the cosine distance $\psi  = 1 - \cos(\mathbf{u}_{w},\mathbf{v}_{w})$.
The vector $\mathbf{v}_{w}$ is generated by the following equation:
\begin{equation}
\mathbf{v}_{w} = \phi  ( \zeta(w) ), \text{ for } w \in \mathcal{V} \text{ or } w \notin \mathcal{V}
\end{equation}
\noindent Here, $\zeta(\cdot)$ is a function that maps $w$ to a list of subunits based on the surface form of the word (e.g., a character or subword sequence).
After that, the sequence is fed into the function $\phi(\cdot)$ to produce vectors, and the inside structure can be CNNs, RNNs, or a simple summation function.
After training, the model can impute vectors for arbitrary words.
Table~\ref{tab:mimic-like table} shows some features of three mimick-like models.
\lihu{done}

\subsection{Contrastive Learning}
\ignore{
Mimick-like models do not require re-training on large-scale data and can efficiently generate vectors for unseen words.
However, there are two main drawbacks in prior methods:
(1) In general, these models adopt mean squared loss to minimize the distance between the generated and target vectors (positive pairs)  while ignoring to maximize the distance between negative pairs, especially for the word pairs that look very alike but with totally different meanings. 
(2) These models cannot produce consistently robust representations for words with small character perturbations.
}

Contrastive learning methods have achieved significant success for image representation~\cite{oord2018representation,chen2020simple}. 
The core idea of these methods is to encourage learned representations for positive pairs to be close, while pushing representations from sampled negative pairs apart. The widely used contrastive loss~\cite{hjelm2018learning,logeswaran2018an, chen2020simple,he2020momentum} is defined as:
\begin{align}\label{eq:cl_loss}
% \mathcal{L}_{cl} = - \log \frac{e^{(\mathbf{u}_{i}^{\mathsf{T}}\mathbf{u}^{+}/ \tau)} }{\sum_{j=1}^{N} e^{(\mathbf{u}_{i}^{\mathsf{T}}\mathbf{u}_{j}/ \tau)}}
\ell_{cl} &= - \log \frac{e^{\text{sim}(\mathbf{u}_{i}^{\mathsf{T}}\mathbf{u}^{+})/ \tau} }{ e^{\text{sim}(\mathbf{u}_{i}^{\mathsf{T}}\mathbf{u}^{+})/ \tau} + \sum e^{\text{sim}(\mathbf{u}_{i}^{\mathsf{T}}\mathbf{u}^{-})/ \tau}}
\end{align}
\noindent Here, $\tau $ is a temperature parameter, $\text{sim}(\cdot)$ is a similarity function such as cosine similarity, and $(u_{i},u^{+})$, $(u_{i},u^{-})$ are positive pairs and negative pairs, respectively (assuming that all vectors are normalized).
During training, positive pairs are usually obtained by augmentation for the same sample, and negative examples are the other samples in the mini-batch. 
This process learns representations that are invariant against noisy factors to some extent.

\begin{figure*}[t]
	\centering
	\includegraphics[width=1\textwidth]{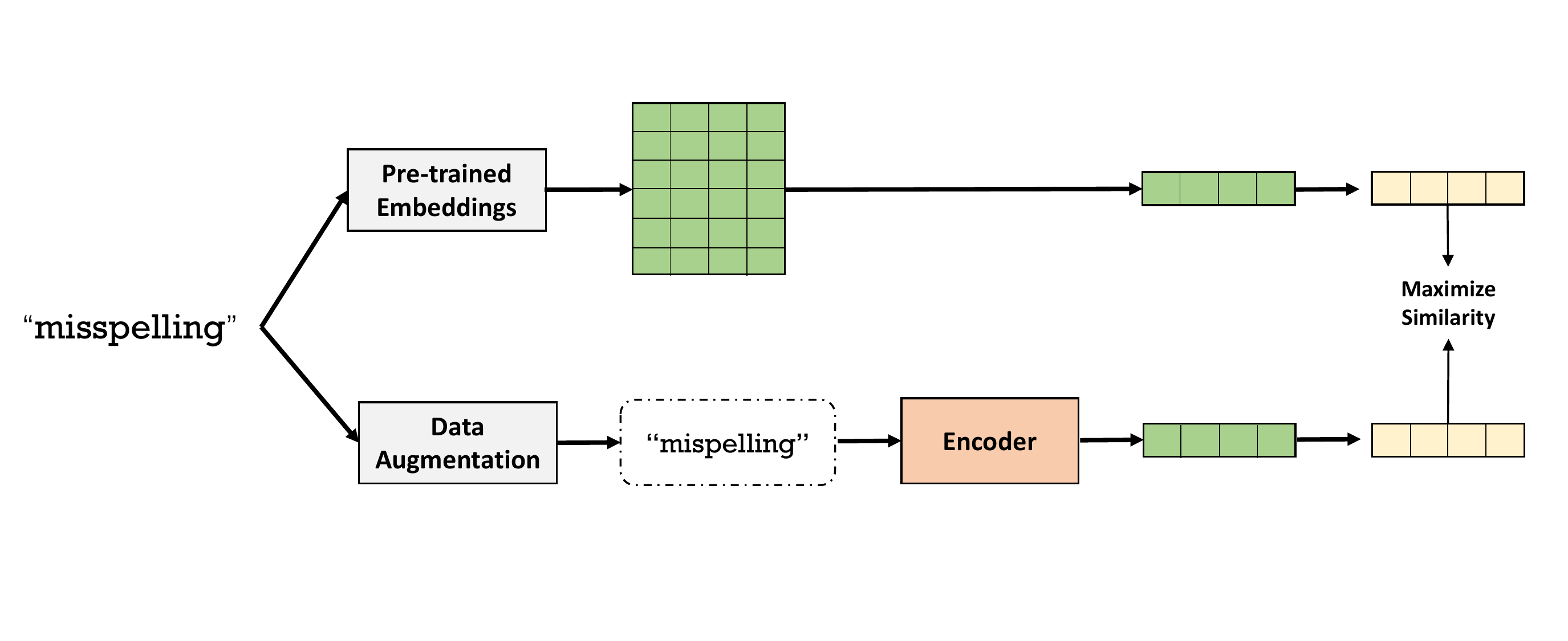}
	\caption{The framework of LOVE with an example of the word \texttt{misspelling}. }
	\label{fig:clearning}
\end{figure*}

\section{Our Approach: LOVE} \label{sec:approach}

LOVE (Learning Out-of-Vocabulary Embeddings) draws on the principles of contrastive learning to maximize the similarity between target and generated vectors, and to push apart negative pairs. An overview of our framework is shown in Figure~\ref{fig:clearning}. It is inspired by work in visual representation learning~\cite{chen2020simple}, but differs in that one of the positive pairs is obtained from pre-trained embeddings instead of using two augmented versions. 
We adopt five novel types of word-level augmentations and a lightweight Positional Attention Module in this framework.
Moreover, we find that adding hard negatives during training can effectively yield better representations.
We removed the nonlinear projection head after the encoder layer, 
%that can improve the representation quality in the visual field,
because its improvements are specific to the representation quality in the visual field. 
%we do not see clear benefit of this strategy in our setting.
Furthermore, our approach is not an unsupervised contrastive learning framework, but a supervised learning approach.

Our framework takes a word from the original vocabulary and uses data augmentation to produce a corruption of it. For example,  \texttt{"misspelling"} becomes \texttt{"mispelling"} after dropping one letter \texttt{"s"}. 
Next, we obtain a target vector from the pre-trained embeddings for the original word, and we generate a vector for the corrupted word.
These two vectors are a pair of positive samples, and we maximize the similarity between them while making the distance of negative pairs (other samples in the same mini-batch) as large as possible.
As mentioned before, we use the contrastive loss as an objective function (Eq~\ref{eq:cl_loss}).
There are five key ingredients in the framework that we will detail in the following (similar to the ones in Table~\ref{tab:mimic-like table}): the Input Method, the Encoder, the Loss Function, our Data Augmentation, and the choice of Hard Negatives.

\subsection{Input Method}
Our goal is to use the surface form to impute vectors for words.
The question is thus how to design the function $\zeta(\cdot)$ mentioned in Section~\ref{sec:mimic_like_model} to represent each input word. 
MIMICK~\cite{pinter2017mimicking} straightforwardly uses the character sequence (see Table~\ref{tab:mimic-like table}). This, however, loses the information of morphemes, i.e., sequences of characters that together contribute a meaning. Hence, FastText~\cite{bojanowski2017enriching} adopts character n-grams. 
Such n-grams, however, are highly redundant.
For example, if we use substrings of length 3 to 5 to represent the word \texttt{misspelling}, we obtain a list with 24 n-gram characters -- while only the substrings \texttt{\{mis, spell, ing\}} are the three crucial units to understand the word.  
Hence, like BERT, we use WordPiece~\cite{wu2016google} with a vocabulary size of around 30000 to obtain meaningful subwords of the input word.
For the word \texttt{misspelling}, this yields \{\texttt{miss}, \texttt{\#\#pel}, \texttt{\#\#ling} \}.
However, if we just swap two letters (as by a typo), then the sequence becomes completely different: \{\texttt{mi}, \texttt{\#\#sp}, \texttt{\#\#sell}, \texttt{\#\#ing} \}.
Therefore, we use both the character sequence and subwords  (Figure~\ref{fig:mixed_input}).

We shrink our vocabulary by stemming all words and keeping only the base form of each word, and by removing words with numerals. This decreases the size of vocabulary from 30\,000 to 21\,257 without degrading performance too much (Section~\ref{sec:shrinking}).

\ignore{
To solve this problem, we combine character and subword sequences to represent each word, as a \textit{Mixed input}.
As BERT~\cite{devlin2019bert}, we adopt WordPiece~\cite{wu2016google} with a 30,000 token vocabulary to obtain subwords.
The beginning of a sequence is always a special token \texttt{[CLS]}, and next is the character sequence.
We insert a special token \texttt{[SUB]} to connect characters and subwords. At the end of each sequence is the token \texttt{[SEP]}.\fms{How does this solve the problem? If this is the same as what BERT does, no need to elaborate on n-grams etc.}
}

\subsection{Encoder}
Let us now design the function $\phi(\cdot)$ mentioned in Section~\ref{sec:mimic_like_model}. 
We are looking for a function that can encode both local features and global features.
Local features are character n-grams, which provide robustness against minor variations such as character swaps or omissions.
Global features combine local features regardless of their distance. 
For the word \texttt{misspelling}, a pattern of prefix and suffix \texttt{mis}+\texttt{ing} can be obtained by combining the local information at the beginning and the end of the word. 
Conventional CNNs, RNNs, and self-attention cannot extract such local and global information at the same time.
Therefore, we design a new \textbf{Positional Attention Module}. 
Suppose we have an aforementioned mixed input sequence and a corresponding embedding matrix $\mathbf{V} \in \mathbb{R}^{ |\mathcal{V}| \times d}$ where $d$ is the dimension of vectors. 
Then the input can be represented by a list of vectors: $\mathbf{X}=\{\mathbf{x}_{1}, \mathbf{x}_{2},...,
\mathbf{x}_{n}\} \in \mathbb{R}^{n \times d}$ where $n$ is the length of the input. 
To extract local information, we first adopt positional attention to obtain n-gram features, and then feed them into a conventional self-attention layer to combine them in a global way. 
This can be written as:
\begin{align}
\label{eq:sa_and_pa}
\mathbf{\bar{X}} &=  \mathrm{SA}(\mathrm{PA}(\mathbf{X})) \, \mathbf{W}^{O} 
\end{align}
\noindent Here, SA is a standard multi-head self-attention and PA is a positional attention.
$\mathbf{W}^{O} \in \mathbb{R}^{ d_{V} \times d_{O}}$ is a trainable parameter matrix, where $d_{V}$ are the dimensions of values in SA and PA, and $d_{O}$ is that of $\mathbf{\bar{X}}$.
As for the Positional Attention, we adopt absolute sinusoidal embeddings~\cite{vaswani2017attention} to compute positional correlations:
\begin{align}
	\mathrm{PA}(\mathbf{X}) =
	\mathrm{Softmax}\biggl(\frac{\mathbf{P}\mathbf{P}^{\mathsf{T}}}{\sqrt{d}}\biggr)(\mathbf{X\,W}^{V}) 
\end{align}
\noindent Here, $\mathbf{P} \in \mathbb{R}^{ n \times d}$ are the position embeddings, and $\mathbf{W}^{V} \in \mathbb{R}^{ d \times d_{V}}$ are the corresponding parameters.
More details about the encoder are in Appendix~\ref{sec:app_encoder}.

\subsection{Loss Function}\label{sec:loss}
In this section, we focus on the loss function $\mathcal{L}(\cdot)$.
Mimick-like models often adopt the mean squared error (MSE), which tries to give words with the same surface forms similar embeddings.
However, the MSE only pulls positive word pairs closer, and does not push negative word pairs apart.
Therefore, we use the contrastive loss instead (Equation~\ref{eq:cl_loss}). \citet{wang2020understanding} found that the contrastive loss optimizes two key properties: \textit{Alignment} and \textit{Uniformity}.
The Alignment describes the expected distance (closeness) between positive pairs:

\begin{equation}
\ell_{\text{align}} \triangleq \mathop{\mathbb{E}}\limits_{(x,y) \sim p_{\text{pos}}} {\psi (\mathbf{u}_{x}, \mathbf{u}_{y})}
\end{equation}
Here, $p_{\text{pos}}$ is the distribution of positive pairs. 
The Uniformity measures whether the learned representations are uniformly distributed in the hypersphere:
\begin{equation}
\ell_{\text{uniform}} \triangleq \log \mathop{\mathbb{E}}\limits_{(x,y) \overset{i.i.d.}\sim\ p_{\text{data}}} {e^{-t \cdot \psi (\mathbf{u}_{x}, \mathbf{u}_{y})}}
\end{equation}
Here, $p_{\text{data}}$ is the data distribution and $t>0$ is a parameter.
The two properties are consistent with our expected word representations: positive word pairs should be kept close and negative word pairs should be far from each other, finally scattered over the hypersphere.

\ignore{
To solve this problem, it is natural to introduce the contrastive loss~\cite{hjelm2018learning,logeswaran2018an, chen2020simple,he2020momentum}:

\begin{equation}\label{eq:cl_loss1}
\mathcal{L}_{cl} = - \log \frac{e^{(\mathbf{u}_{i}^{\mathsf{T}}\mathbf{u}^{+}/ \tau)} }{\sum_{j=1}^{N} e^{(\mathbf{u}_{i}^{\mathsf{T}}\mathbf{u}_{j}/ \tau)}}
\end{equation}
where $\tau $ is a temperature parameter, $N$ is the batch size and $(u_{i},u^{+})$ is a positive pair (assume all vectors are normalized).
Following the work of~\citet{wang2020understanding}, the above loss function can be rewrote as:

\begin{align} 
\mathcal{L}_{cl} &= - \log \frac{e^{(\mathbf{u}_{i}^{\mathsf{T}}\mathbf{u}^{+}/ \tau)} }{ e^{(\mathbf{u}_{i}^{\mathsf{T}}\mathbf{u}^{+}/ \tau)} + \sum e^{(\mathbf{u}_{i}^{\mathsf{T}}\mathbf{u}^{-}/ \tau)}} & \\
&=  -\mathbf{u}_{i}^{\mathsf{T}}\mathbf{u}^{+}/ \tau \notag
+ \log \sum e^{(\mathbf{u}_{i}^{\mathsf{T}}\mathbf{u}^{-}/ \tau)} 
\\ \notag
\end{align}
where the first part corresponds to the optimization of alignment metric, and the second part corresponds to the Uniformity. You can find more formal proofs in the original work.
}

\subsection{Data Augmentation and Hard Negatives}\label{sec:data_aug}

\begin{figure}[t]
	\centering
	\includegraphics[width=0.5\textwidth]{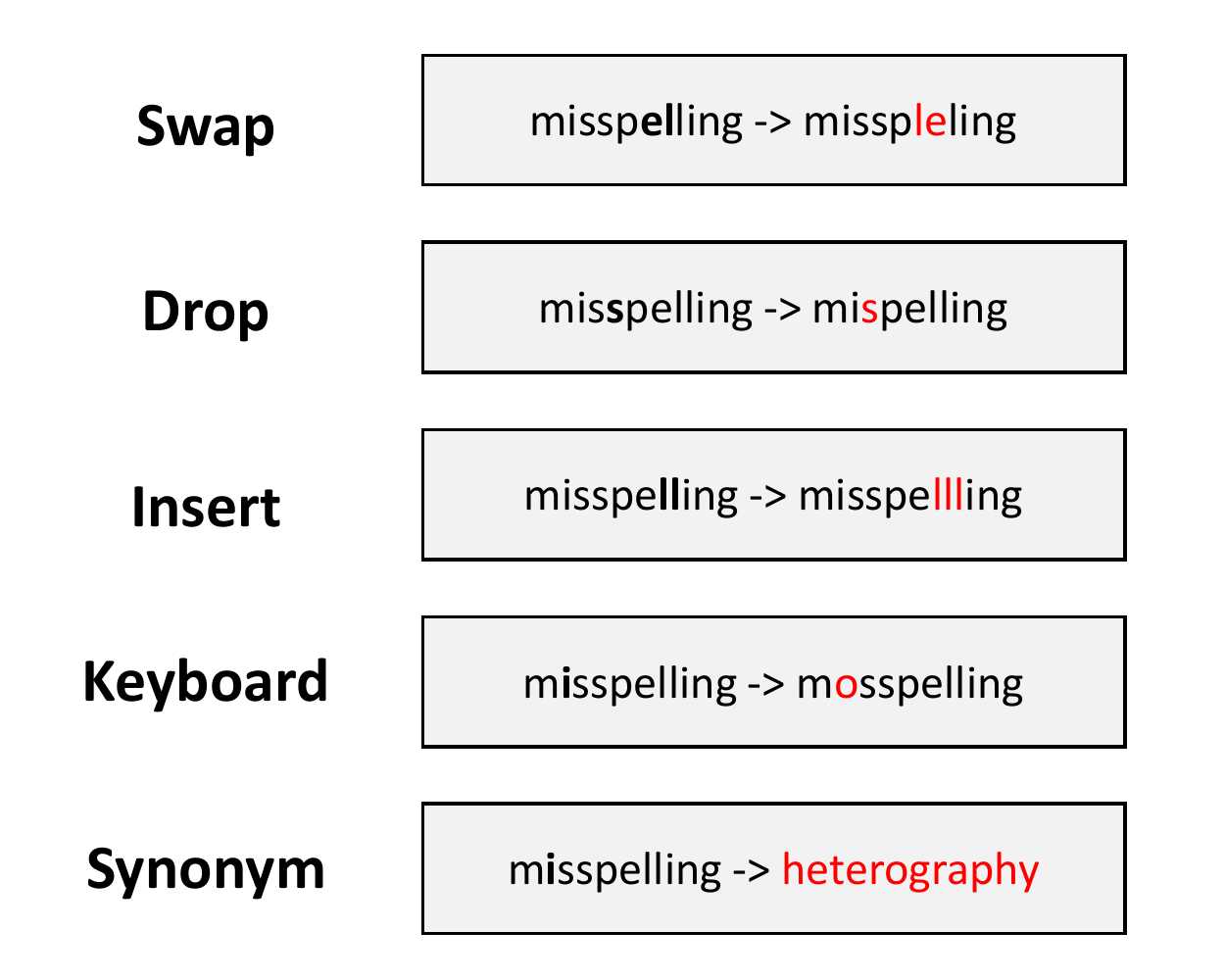}
	\caption{Illustrations of different augmentations for the word \texttt{misspelling}.}
	\label{fig:augmentation}
\end{figure}

Our positive word pairs are generated by data augmentation, which can increase the amount of training samples by using existing data.
We use various strategies (Figure ~\ref{fig:augmentation}) to increase the diversity of our training samples: (1) Swap two adjacent characters,
(2) Drop a character, 
(3) Insert a new character, 
(4) Replace a character according to keyboard distance,
(5) Replace the original word by a synonymous word.
The first four augmentations are originally designed to protect against adversarial attacks~\cite{pruthi2019combating}. We add the synonym replacement strategy to keep semantically similar words close in the embedding space --  something that cannot be achieved by the surface form alone. 
Specifically, a set of synonyms is obtained by retrieving the nearest neighbors from pre-trained embeddings like FastText. 

%The choice of negative pairs impacts the quality of learned features in the framework a lot, and the current negatives are randomly sampled from the dataset.
Negative word pairs are usually chosen randomly from the mini-batch. However, we train our model to be specifically resilient to \emph{hard negatives (or difficult negatives)}, i.e., words with similar surface forms but different meanings (e.g., \textit{misspelling} and \textit{dispelling}).
%To make our model distinguish hard negatives (words with alike surface form but different meanings\fms{important novel concept!}),  we select the most indistinguishable samples as hard negative samples.\lihu{And this procedure is important.}
To this end, we add a certain number of hard negative samples (currently 3 of them) to the mini-batch, by selecting word pairs that are not synonyms and have a small edit distance.

\ignore{
\subsection{Shrinking Our Model}

\textcolor{blue}{We shrink our vocabulary by stemming all words and keeping only the base form of each word, and by removing words with numerals. This decreases the number of parameters from 9M to 6.3M without degrading performance too much.}
\textcolor{gray}{
In our model, the subword embedding matrix $\mathbf{V} \in \mathbb{R}^{ |\mathcal{V}| \times m}$ that we need to learn occupies a large proportion of the total parameters.
To simplify our model, we can either reduce the size of the predefined vocabulary $\mathcal{V}$ or reduce the dimension of the subword vector $m$.
Here, we find it is very effective to reduce the size of the vocabulary $\mathcal{V}$.
%In detail, the original vocabulary contains plurals and conjugations, therefore 
We stem all complete words and remove words with numerals.
Finally, we obtain a new vocabulary with $21257$ words, and the embedding parameters shrank from 9M to 6.3M without degrading the performance too much.}
}

\subsection{Mimicking Dynamical Embeddings} \label{sec:mimic_dynamics}
Pre-trained Language Models (e.g., ELMo~\cite{peters-etal-2018-deep} and BERT~\cite{devlin2019bert}) dynamically generate word representations based on specific contexts, which cannot be mimicked directly.
To this end, we have two options: We can either learn the behavior of the \emph{input embeddings} in BERT before the multi-layer attentions or mimic the static \emph{distilled embeddings}~\cite{bommasani2020interpreting,gupta2021obtaining}.

We use the BERT as an example to explain these two methods. Suppose we have a subword sequence after applying WordPiece to a sentence: $W = \{w_{1}, w_{2}, ..., w_{n} \}$.
For the subword sequence $W$, BERT first represents it as a list of subword embeddings: $\mathbf{E}^{in} = \{\mathbf{e}^{sub}_{1}, \mathbf{e}^{sub}_{2}, ..., \mathbf{e}^{sub}_{n} \}$. 
We refer to this static representation as the Input Embedding of BERT, and we can use our model to mimic the behavior of this part. We call this method \emph{mimicking input embeddings}.
For ease of implementation, we learn only from the words that are not separated into pieces.
After that step, BERT applies a multi-layer multi-head attention to the input embeddings $E^{in}$, which yields a contextual representation for each subword: $\mathbf{E}^{out} = \{\mathbf{e}^{out}_{1}, \mathbf{e}^{out}_{2}, ..., \mathbf{e}^{out}_{n} \}$. 
However, these contextual representations vary according to the input sentence and we cannot learn from them directly.
Instead, we choose to mimic the distilled static embeddings from BERT, which are obtained by pooling (max or average) the contextual embeddings of the word in different sentences. We call this method \emph{mimicking distilled embeddings}.
The latter allows for better word representations, while the former does not require training on a large-scale corpus. Our empirical studies show that mimicking distilled embeddings performs only marginally better. 
Therefore, we decided to rather learn the input embeddings of BERT, which is simple yet effective

\subsection{Plug and Play}
One of the key advantages of our model is that it can be used as a plug-in for other models. For models with static word embeddings like FastText, one can simply use our model to generate vectors for unseen words. For models with dynamic word embeddings like BERT,
if a single word is tokenized into several parts, e.g. \texttt{misspelling} = \{\texttt{miss}, \texttt{\#\#pel}, \texttt{\#\#ling} \}, we regard it as an OOV word.
Then, we replace the embeddings of the subwords by a single embedding produced by our model before the attention layer. Our final enhanced BERT model has 768 dimensions and 16M parameters. Note that the BERT-base model has \textasciitilde110M parameters and its distilled one has \textasciitilde 550M parameters.

\ignore{
we have two options: We can either mimic the static \emph{distilled embeddings}~\cite{bommasani2020interpreting,gupta2021obtaining} or learn the behavior of the \emph{input embeddings} in BERT before the multi-layer attentions.
The former allows for better word representations, while the latter does not require training on a large-scale corpus. Our empirical studies (Section~\ref{sec:work_with_bert}) show that mimicking distilled embeddings performs only marginally better. 
Therefore, we decided to rather learn the input embeddings of BERT, which is simple yet effective. 
Since BERT can cope with unseen words through WordPiece, we define an OOV word in BERT as a word that is separated into several pieces, e.g. \texttt{misspelling} = \{\texttt{miss}, \texttt{\#\#pel}, \texttt{\#\#ling} \}.
}

\begin{table*}[hbt] 
	\centering 
	\scriptsize
	\begin{threeparttable} 
		\begin{tabular}{cccccccccccc}  
			\toprule  &\multicolumn{2}{c}{parameters}&\multicolumn{6}{c}{Word Similarity}&\multicolumn{2}{c}{Word Cluster}&Avg\cr
			&embedding&others&RareWord&SimLex&MTurk&MEN&WordSim&SimVerb&AP&BLESS\cr
			\midrule
			FastText \shortcite{bojanowski2017enriching} &969M&-&48.1&30.4&66.9&78.1&68.2&25.7&58.0&71.5&55.9\cr
			\midrule
			MIMICK  \shortcite{pinter2017mimicking} &9M&517K&27.1&15.9&32.5&36.5&15.0&7.5&\textbf{59.3}&\textbf{72.0}&33.2\cr
			BoS  \shortcite{zhao2018generalizing} &500M&-&\textbf{44.2}&\underline{27.4}&\underline{55.8}&\underline{65.5}&\underline{53.8}&\underline{22.1}&41.8&39.0&\underline{43.7}\cr
			%KVQ-FH  \shortcite{sasaki2019subword} &9M&-&\textbf{49.8}&31.9&\textbf{67.3}\cr
			KVQ-FH  \shortcite{sasaki2019subword} &12M&-&\underline{42.4}&20.4&55.2&63.4&53.1&16.4&39.1&42.5&41.6\cr
			LOVE&6.3M&200K&42.2&\textbf{35.0}&\textbf{62.0}&\textbf{68.8}&\textbf{55.1}&\textbf{29.4}&\underline{53.2}&\underline{51.5}&\textbf{49.7}\cr
			%Ours-big &9M&200K&40.4&\textbf{32.6}&\textbf{59.9}&\textbf{65.7}&\textbf{56.0}&\underline{22.0}&\underline{51.0}&\underline{60.5}&\textbf{48.5}\cr
			\bottomrule  
		\end{tabular}
		\caption{Performance on the intrinsic tasks, measured as Spearman’s $\rho$ and purity for word similarity and clustering. Best performance among the mimick-like models in bold, second-best underlined.}	
		\label{tab:intrinsic_results}
	\end{threeparttable} 
	%\end{minipage}%
\end{table*}
\begin{table*}[hbt] 
	\centering 
	\scriptsize
	\begin{threeparttable} 
		\begin{tabular}{cccccccccccc}  
			\toprule  &\multicolumn{2}{c}{parameters}&\multicolumn{2}{c}{\textbf{SST2}}&\multicolumn{2}{c}{\textbf{MR}}&\multicolumn{2}{c}{\textbf{CoNLL-03}}&\multicolumn{2}{c}{\textbf{BC2GM}}&Avg\cr
			&embedding&others&original&+typo&original&+typo&original&+typo&original&+typo&\cr
			\midrule
			FastText \shortcite{bojanowski2017enriching} &969M&-&82.3&60.5&73.3&62.2&86.4&66.3&71.8&53.4&69.5\cr
			Edit Distance  &969M&-&-&67.4&-&68.3&-&76.2&-&66.6&-\cr
			\midrule
			MIMICK  \shortcite{zhao2018generalizing} &9M&517K&69.7&62.3&\underline{73.6}&61.4&68.0&65.2&56.6&56.7&64.2\cr
			BoS  \shortcite{zhao2018generalizing} &500M&-&\underline{79.7}&\underline{72.6}&\underline{73.6}&\textbf{69.5}&\textbf{79.5}&68.6&\textbf{66.4}&\underline{61.5}&\underline{71.5}\cr
			KVQ-FH  \shortcite{sasaki2019subword} &12M&-&77.8&71.4&72.9&66.5&73.1&\textbf{70.4}&46.2&53.5&66.5\cr
			%Ours &6.3M&200K&\underline{80.7}&\textbf{73.7}&73.3&67.7&78.2&67.8&63.8&\textbf{63.8}&71.1\cr
			LOVE &6.3M&200K&\textbf{81.4}&\textbf{73.2}&\textbf{74.4}&\underline{66.7}&\underline{78.6}&\underline{69.7}&\underline{64.7}&\textbf{63.8}&\textbf{71.6}\cr
			%Ours-big &9M&200K&\textbf{81.3}&72.5&\textbf{74.7}&\underline{68.2}&\textbf{80.2}&\underline{69.0}&\underline{66.0}&\underline{61.5}&\textbf{71.7}\cr
			%Corrector  &-&-&-&75.3&-&69.7&-&70.4&-&-&-\cr
			\bottomrule  
		\end{tabular}
		\caption{Performance on the extrinsic tasks, measured as accuracy and F1 (five runs of different learning rates) for text classification and NER, respectively. Typos are generated by simulated errors of an OCR engine~\cite{ma2019nlpaug}. The speed of producing word vectors with Edit Distance and LOVE is \emph{380s/10K} words and \emph{0.9s/10K} words, respectively.}	
		\label{tab:extrinsic_results}
	\end{threeparttable} 
	%\end{minipage}%
\end{table*}

\section{Experiments}\label{sec:experiments}
\subsection{Evaluation Datasets}
There are two main methods to evaluate word representations: Intrinsic and Extrinsic.
Intrinsic evaluations measure syntactic or semantic relationships between words directly, e.g., word similarity in word clusters.
Extrinsic evaluations measure the performance of word embeddings as input features to a downstream task, e.g., named entity recognition (NER) and text classification.
Several studies have shown that there is no consistent correlation between intrinsic and extrinsic evaluation results \cite{chiu2016intrinsic,faruqui2016problems,wang2019evaluating}.
Hence, we evaluate our representation by both intrinsic and extrinsic metrics.
Specifically, we use 8 intrinsic datasets (6 word similarity and 2 word cluster tasks):
RareWord~\cite{luong2013better},
SimLex~\cite{hill2015simlex},
MTurk~\cite{halawi2012large},
MEN~\cite{bruni2014multimodal},
WordSim~\cite{agirre2009study},
Simverb~\cite{agirre2009study},
AP~\cite{almuhareb2006attributes} and 
BLESS~\cite{baroni2011we}.
We use four extrinsic datasets (2 text classification and 2 NER tasks):
SST2~\cite{socher2013recursive}, MR~\cite{pang2005seeing},
CoNLL-03~\cite{sang2003introduction} and BC2GM~\cite{smith2008overview}.
It is worth noting that the RareWord dataset contains many long-tail words and the BC2GM is a domain-specific NER dataset.
All data augmentations and typo simulations are implemented by NLPAUG\footnote{\url{https://github.com/makcedward/nlpaug}}.
Appendix~\ref{sec:detail_experiment} provides more details on our datasets and experimental settings.

\ignore{
\paragraph{Training Details}
The target pre-trained embeddings are FastText \cite{bojanowski2017enriching} because it is a strong baseline by summing up subword-level information to produce vectors for arbitrary words.
Besides, we use MIMICK, BoS, and KVQ-FH that do not train on contextual words as competitors.
Note that we do not adopt Robust Backed-off Estimation \cite{fukuda2020robust} and PBoS \cite{jinman2020pbos} as baselines as the two  models make use of larger and more complex models to obtain word representations.
Robust Backed-off Estimation uses string matching to find the top-k similar words from the entire vocabulary when imputing.
Using the same target vectors, the parameter numbers of BoS and PBoS are 163M and 316M respectively.
Therefore we choose the three baselines and follow the official setting to train every MIMICK-like model on \texttt{fasttext-crawl-300d-2M}\footnote{\url{https://fasttext.cc/docs/en/english-vectors.html}}.

We choose the setting discussed in Section~\ref{sec:approach} to train our model for 20 epochs, and evaluate each intrinsic task based on the produced vectors of models.
As for extrinsic tasks, we feed word vectors to each neural network and fix them during training.
We use CNNs for text classification \cite{zhang2015sensitivity} and BiLSTM+CRF for NER \cite{huang2015bidirectional}. 
Then, we report the best result among five different learning rates (see Appendix~\ref{sec:detail_experiment} for more details). 
}

\subsection{Results on Intrinsic Tasks}
Table~\ref{tab:intrinsic_results} shows the experimental results on 8 intrinsic tasks. Compared to other mimick-like models, our model achieves the best average score across 8 datasets while using the least number of parameters.
Specifically, our model performs best on 5 word similarity tasks, and second-best % while only worse than the adapted MIMICK 
on the word cluster tasks.
Although there is a gap between our model and the original FastText, we find our performance acceptable, given that our model is 100x times smaller.

\begin{table*}[hbt] 
	\centering 
	\scriptsize
	\begin{threeparttable} 
		\begin{tabular}{cccccccccccccc}  
			\toprule  &\multicolumn{6}{c}{\textbf{SST2}}&\multicolumn{6}{c}{\textbf{CoNLL-03}}&\cr
			Typo Probability&original&10\%&30\%&50\%&70\%&90\%&original&10\%&30\%&50\%&70\%&90\%&Avg\cr
			\midrule
			\multicolumn{13}{c}{Static Embeddings}\cr
			\midrule
			FastText  &\textbf{82.3}&68.2&59.8&56.7&57.8&60.3&\textbf{86.4}&81.6&78.9&73.9&70.2&63.4&70.0\cr
			FastText + LOVE&82.1&\textbf{79.8}&\textbf{74.9}&\textbf{74.2}&\textbf{68.8}&\textbf{67.2}&86.3&\textbf{84.7}&\textbf{81.8}&\textbf{77.5}&\textbf{73.1}&\textbf{71.3}&\textbf{76.8}\cr
			\midrule
			\multicolumn{13}{c}{Dynamical Embeddings}\cr
			\midrule
			BERT  &\textbf{91.5}&88.2&78.9&74.7&69.0&60.1&\textbf{91.2}&\textbf{89.8}&\textbf{86.2}&83.4&79.9&76.5&80.7\cr
			BERT + LOVE &\textbf{91.5}&\textbf{88.3}&\textbf{83.7}&\textbf{77.4}&\textbf{72.7}&\textbf{63.3}&89.9&88.3&86.1&\textbf{84.3}&\textbf{80.8}&\textbf{78.3}&\textbf{82.1}\cr
			\bottomrule  
		\end{tabular}
		\caption{Robust evaluation (five runs of different learning rates) on text classification and NER under simulated post-OCR typos. We use uncased and cased BERT-base model for SST2 and CoNLL-03, respectively.}	
		\label{tab:typo_check}
	\end{threeparttable} 
	%\end{minipage}%
\end{table*}

\subsection{Results on Extrinsic Tasks}
Table~\ref{tab:extrinsic_results} shows the results on four downstream datasets and their corrupted version.
In this experiment, we introduce another non-trivial baseline: Edit Distance.
For each corrupted word, we find the most similar word from a vocabulary using edit distance and then use the pre-trained vectors of the retrieved word.
Considering the time cost, we only use the first 20K words appearing in FastText (2M words) as reference vocabulary.

The typo words are generated by simulating post-OCR errors. 
For the original datasets, our model obtains the best results across 2 datasets and the second-best on NER datasets compared to other mimick-like models.
For the corrupted datasets, the performance of the FastText model decreases a lot and 
our model is the second best but has very close scores with BoS consistently.
Compared to other mimick-like models, our model with 6.5M achieves the best average score.
Although Edit Distance can effectively restore the original meaning of word, it is 400x times more time-consuming 
%(without parallelism and the use of indexing methods) 
than our model. 
% Fabian: there is work that can speed up Edit Distance \url{https://scholar.google.com/scholar?q=search+edit+distance}
% ...but that would bring us too far. Anyway, people are not likely to know or care... :-(

\begin{figure}[t]
	\centering
	\includegraphics[width=0.5\textwidth]{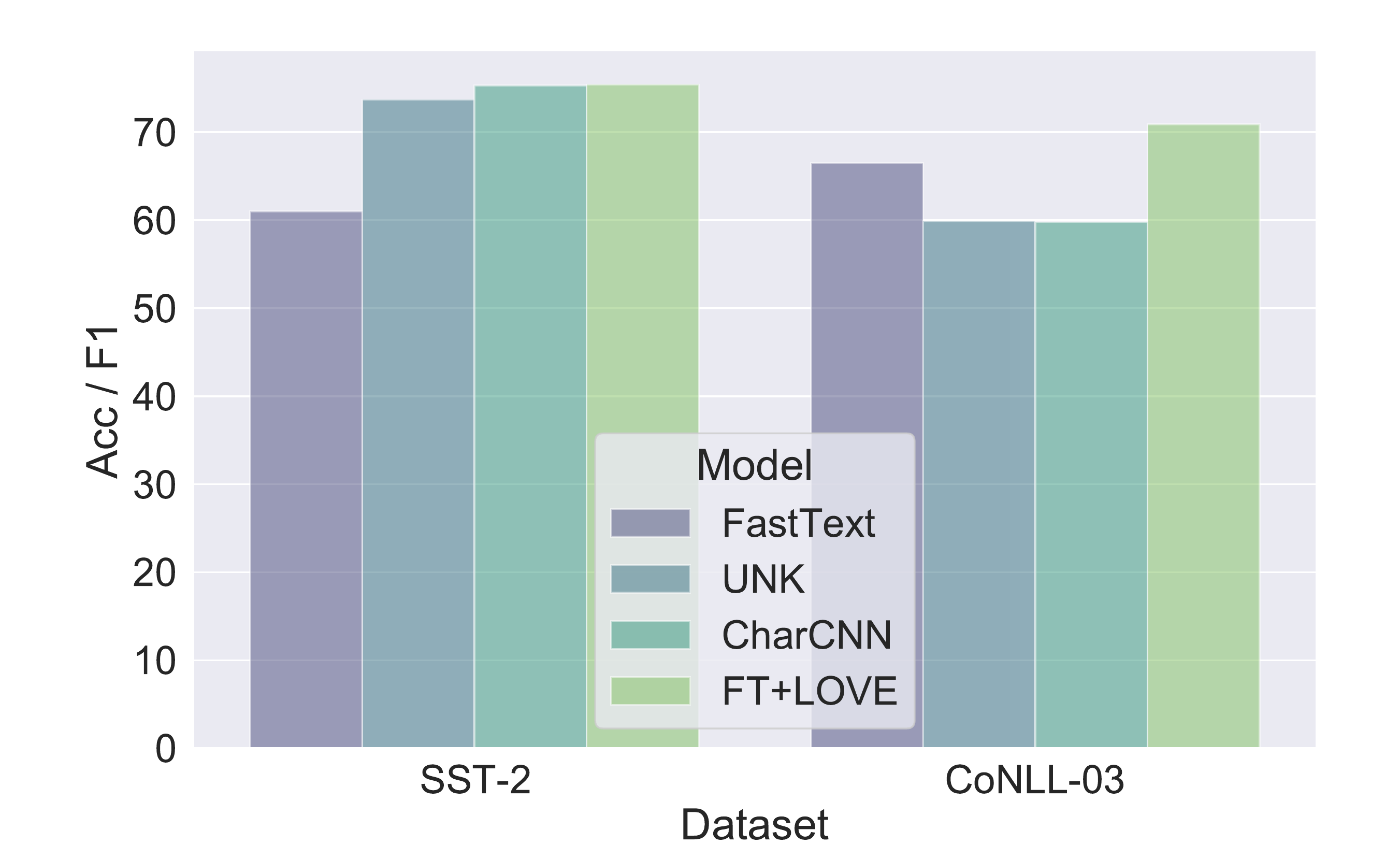}
	\caption{Evaluation of different methods based on FastText under typos. }
	\label{fig:work_with_ft}
\end{figure}

\subsection{Robustness Evaluation}
In this experiment, we evaluate the robustness of our model by gradually adding simulated post-OCR typos~\cite{ma2019nlpaug}.
Table~\ref{tab:typo_check} shows the performances on SST2 and CoNLL-03 datasets.
We observe that our model can improve the robustness of the original embeddings without degrading their performance.
Moreover, we find our model can make FastText more robust compared to other commonly used methods against unseen words:
a generic \texttt{UNK} token or character-level representation of neural networks. 
Figure~\ref{fig:work_with_ft} shows the robustness check of different strategies.
FastText+LOVE has a consistent improvement on both SST2 and CoNLL-03 datasets. At the same time, LOVE degrades the performance on the original datasets only marginally if at all.

\ignore{
\paragraph{Working with FastText Together}
To solve the OOV problem, conventional methods use a generic \texttt{UNK} token or character-level representation of neural networks. 
Actually, our model can mimic the behavior of well-trained embeddings and produce vectors for OOV words.
Therefore, our model can work with the target model as a supplement for unseen words.
In this experiment, we empirically demonstrate our model can make FastText more robust compared to other common-used methods.
Figure~\ref{fig:work_with_ft} shows the robustness check of different strategies.
On text classification dataset, \texttt{UNK} and CharCNN can substantially improve the robustness compared to the subword model FastText, and our FT+Ours only has a small advantage.
However, FT+Ours has a clear lead of at least five points on token-level tasks under typos. 
}
\subsection{Ablation Study}
We now vary the components in our architecture (input method, encoder and loss function) to demonstrate the effectiveness of our architecture.
%We evaluate various methods on RareWord set \cite{luong2013better} and SST2 dataset\cite{socher2013recursive} (one for intrinsic evaluation and one for extrinsic evaluation).

\paragraph{Input Method.}
To validate the effect of our Mixed Input strategy, we compare it with two other methods:
using only the character sequence or only the subword sequence.
Table~\ref{tab:architecure_analysis} shows that the Mixed method achieves better representations, and any removal of char or subword information can decrease the performance.

\begin{table}[hbt] 
	\centering 
	\scriptsize
	\begin{threeparttable} 
		\begin{tabular}{ccccc}  
			\toprule
			&\multicolumn{2}{c}{parameters}&RareWord&SST2\cr
			&\!\!\!embedding&others&\cr
			\midrule
			The original LOVE &6.3M&200K&42.2&81.4\cr
			\midrule
			\multicolumn{5}{c}{Varying the input method}\cr
			\midrule
            only use Char & 299K&200K&17.7&71.5\cr
            only use Subword &6.0M&200K&25.3&76.0\cr
            \midrule
            \multicolumn{5}{c}{Varying the encoder}\cr
            \midrule
			replace PAM with CNN &6.3M&270K&28.4&61.1\cr
			replace PAM with RNN &6.3M&517K&27.2&67.2\cr
            replace PAM with SA &6.3M&-&36.9&78.7\cr
            \midrule
            \multicolumn{5}{c}{Varying the loss function}\cr
            \midrule
            use MSE &6.3M&200K&34.5&76.0\cr
            use $\ell_{au} (\lambda=1.0)$ &6.3M&200K&40.8&80.8\cr
			\midrule
			\multicolumn{5}{c}{Ablation of data augmentation and hard negatives}\cr
			\midrule
			w/out hard negatives & 6.3M&200K&37.7&78.6\cr
			w/out hard negatives & &\\
			$~~~~~~~~~~$and augmentation & 6.3M&200K&37.8&78.2\cr
			\bottomrule  
		\end{tabular}
		\caption{\textbf{Ablation studies for the architecture of LOVE}, measured as Spearman’s $\rho$ and accuracy, respectively. }
		\label{tab:architecure_analysis}
	\end{threeparttable} 
\end{table}

\paragraph{Encoder.}
To encode the input sequence, we developed the Positional Attention Module (PAM), which
first extracts ngram-like local features and then uses self-attention combine them without distance restrictions.
%The PAM can achieve better inductive bias while using less parameters. % Fabian: following sentence says this already.
Table~\ref{tab:architecure_analysis} shows that PAM performs the best, which validates our strategy of incorporating both local and global parts inside a word. % is helpful to capture morphological representations.  
 At the same time, the number of parameters of PAM is acceptable in comparison.
We visualize the attention weights of PAM in Appendix~\ref{sec:app_encoder}, to show how the encoder extracts local and global morphological features of a word.

\paragraph{Loss Function.} LOVE uses the contrastive loss, which increases 
%While mimic-like models are based on the Mean Squared Error, we also explore a loss on 
alignment and uniformity.  
\citet{wang2020understanding} proves that optimizing directly these two metrics leads to comparable or better performance than the original contrastive loss.
Such a loss function can be written as:
\begin{equation}
\ell_{\text{au}} = \ell_{\text{align}} + \lambda \cdot \ell_{\text{uniform}}
\end{equation}
Here, $\lambda$ is a hyperparameter that controls the impact of $\ell_{\text{uniform}}$.
We set this value to 1.0 because it achieves the best average score on RareWord and SST2.  
% \gv{How did you set $\lambda$?} % Fabian: I guess it's just set to 1.
An alternative is to use the Mean Squared Error (MSE), as in mimick-like models.
Table~\ref{tab:architecure_analysis} compares the performances of these different loss functions.
The contrastive loss  significantly outperforms the MSE, and there is no obvious improvement by directly using alignment and uniformity.
%, which requires further exploration. Fabian: let's not invite criticism
We also tried various temperatures $\tau$ for the contrastive loss, and the results are shown in Table~\ref{tab:cl_tau} in the appendix.
In the end, a value of $\tau=0.07$ provides a good performance.
\ignore{which shows a suitable $\tau$ is beneficial for good word representations
because a proper temperature can help the model learn from hard negatives~\cite{chen2020simple}.
\lihu{added an conclusion sentence}
\fms{This just says there is a good tau. But which value is best? Or is it learned?}
% Fabian: better come with a firm conclusion here.
%We find that a suitable $\tau$ is beneficial for good word representations
%because a proper temperature can help the model learn from hard negatives \cite{chen2020simple}.
}

\paragraph{Data Augmentation and Hard Negatives.}
In Table~\ref{tab:architecure_analysis}, we observe that the removal of our hard negatives decreases the performance, which demonstrates the importance of semantically different words with similar surface forms.

\begin{table}[t] 
	\centering 
	\scriptsize
	\begin{threeparttable} 
		\begin{tabular}{cccccc}  
			\toprule  &\multicolumn{4}{c}{SST2}\cr
			typos per sentence  &typo-0&typo-1&typo-2&typo-3\cr
			\midrule
			BERT & \textbf{91.5}& 77.2&73.2&69.4\cr
			\midrule
			Mimicking Input Embeddings\cr
			\midrule
			BERT + Add& 91.3& 77.2&73.5&70.7\cr
			BERT + Linear \shortcite{fukuda2020robust} & 91.4& 79.6&77.2&72.8\cr
			BERT + Replacement & \textbf{91.5}& 81.4&78.7&73.6\cr
			\midrule
			Mimicking Distilled Embeddings\cr
			\midrule
			BERT + Add & 91.3& 78.8&75.6&72.3\cr
			BERT + Linear  \shortcite{fukuda2020robust} & 91.3& 81.4&78.7&73.6\cr
			BERT + Replacement & 91.4& \textbf{81.5}&\textbf{78.9}&\textbf{73.8}\cr
			\bottomrule  
		\end{tabular}
		\caption{Performances of different strategies that work with BERT together, measured as the accuracy among five different learning rates.}	
		\label{tab:bert_robustness}
	\end{threeparttable} 
	%\end{minipage}%
\end{table}

LOVE uses five types of word augmentation.
We find that removing this augmentation does not deteriorate performance too much on the word similarity task, while it causes a 0.4 point drop in the text classification task (the last row in Table~\ref{tab:architecure_analysis}), where data augmentations prove helpful in dealing with misspellings.
We further analyze the performance of single and composite augmentations on RareWord and SST2 in the appendix in Figure~\ref{fig:aug_on_rw} and Figure~\ref{fig:aug_on_sst2}. We find that a combination of all five types yields the best results.
%\lihu{actually, there should have another experiment to check the impact of augmentation on corrupted dataset.}

\subsection{The performance of mimicking BERT}
As described in Section~\ref{sec:mimic_dynamics}, we can mimic the input or distilled embeddings of BERT.
After learning from BERT, we use the vectors generated by LOVE to replace the embeddings of OOV subwords. Finally, these new representations are fed into the multi-layer attentions. We call this method the \emph{Replacement} strategy. 
To valid its effectiveness, we compare it with two other baselines:
\noindent\textbf{(1) Linear Combination}~\cite{fukuda2020robust}. For each subword $\mathbf{e}^{sub}$, the generated vectors of word $\mathbf{e}^{word}$ containing the subwords are added to the subword vectors of BERT:
\begin{align} 
\mathbf{e}^{new} &= (1 - \alpha)~\mathbf{e}^{sub} + \alpha~  \mathbf{e}^{word} & \\
\alpha&=  \text{sigmoid}~(\mathbf{W} \cdot \mathbf{e}^{sub})  \notag
\end{align}
where $\mathbf{e}^{sub} \in \mathbb{R}^{d}$ is a subword vector of BERT, and $\mathbf{e}^{word} \in \mathbb{R}^{d}$ is a generated vector of our model. $\mathbf{W}\in \mathbb{R}^{d}$ are trainable parameters.

\noindent\textbf{(2) Add}. A generated word vector is directly added to a corresponding subword vector of BERT:
\begin{align} 
\mathbf{e}^{new} &=  \mathbf{e}^{sub} +  \mathbf{e}^{word} 
\end{align}

Table~\ref{tab:bert_robustness} shows the result of these strategies. %robustness by using our model.
All of them can bring a certain degree of robustness to BERT without decreasing the original capability, which demonstrates the effectiveness of our framework.
Second, the replacement strategy consistently performs best. 
We conjecture that BERT cannot restore a reasonable meaning for those rare and misspelling words that are tokenized into subwords, and our generated vectors can be located nearby the original word in the space. 
Third, we find mimicking distilled embeddings performs the best while mimicking input embeddings comes close.
Considering that the first method needs training on large-scale data, mimicking the input embeddings is our method of choice.

\section{Conclusion}\label{sec:conclusion}
We have presented a lightweight contrastive-learning framework, LOVE, to learn word representations that are robust even in the face of out-of-vocabulary words.
Through a series of empirical studies, we have shown that our model (with only 6.5M parameters) can achieve similar or even better word embeddings on both intrinsic and extrinsic evaluations compared to other mimick-like models.
Moreover, our model can be added to models with static embeddings (such as FastText) or dynamical embeddings (such as BERT) in a plug-and-play fashion, and bring significant improvements there. For future work, we aim to extend our model to languages other than English.

\section{Acknowledgements}
We sincerely thank all the reviewers for their insightful comments and helpful suggestions.
This work was partially funded by ANR-20-CHIA-0012-01 (“NoRDF”).

\bibliography{custom}
\bibliographystyle{acl_natbib}

\clearpage
\appendix
\setcounter{table}{0}   
\setcounter{figure}{0}
\renewcommand{\thetable}{A\arabic{table}}
\renewcommand{\thefigure}{A\arabic{figure}}
\setcounter{equation}{0}
\renewcommand{\theequation}{A.\arabic{equation}}

\section{Details of Our Approach} \label{sec:app_approach}
\subsection{Shrinking Our Model} \label{sec:shrinking}
We consider the following four methods to reduce the total parameters of our model:

\noindent\textbf{(1) Matrix Decomposition.} The original matrix can be decomposed into two smaller matrices $\mathbf{V} = \mathbf{U} \times \mathbf{M},  \mathbf{U} \in \mathbb{R}^{ |\mathcal{V}| \times h}, \mathbf{M} \in \mathbb{R}^{ h \times m}$ and $h < m$. Here, we set $m=300$ and $h=200$ respectively.  

\noindent\textbf{(2) Top Subword.} We use only the top-k frequent subwords, using the word frequencies from a corpus. We set the parameter $k = 20000$.

\noindent\textbf{(3) Hashing.} We use a hashing strategy to share memory for subwords aiming to reduce the parameters. We use a bucket size of $20000$.

\noindent\textbf{(4) Preprocessing.} The original vocabulary contains plurals and conjugations, therefore we stem all complete words and remove words with numerals, obtaining a new vocabulary of $21257$ words.

Table~\ref{tab:shrink} shows that the preprocessing method can reduce parameters very effectively while obtaining a very competitive performance.

\begin{table}[hbt] 
	\centering 
	\tiny
	\begin{threeparttable} 
		\begin{tabular}{ccccc}  
			\toprule 
			&\multicolumn{2}{c}{parameters}&RareWord&SST2\cr
			 &embedding&non-embedding&\cr
			\midrule
			Original &9M&200K&43.5&80.7\cr
			\midrule
			Decomposition &5.6M&200K&38.1&80.3\cr
			Top-K &6M&200K&39.2&80.1\cr
			Hashing &6M&200K&40.5&80.4\cr
			Preprocessing &6.3M&200K&\textbf{42.4}&\textbf{80.7}\cr
			\bottomrule  
		\end{tabular}
		\caption{Performance of different shrinkage strategies, measured as Spearman’s $\rho$ and accuracy, respectively. The target vectors are from \texttt{fasttext-crawl-300d-2M}.}	
		\label{tab:shrink}
	\end{threeparttable} 
\end{table}

\begin{table}[pt] 
	\centering 
	\scriptsize
	\begin{threeparttable} 
		\begin{tabular}{cccccc}  
			\toprule  Hyperparam  &SST2&MR&CONLL-03&BC2GM\cr
			\midrule
			model & CNN& CNN&BiLSTM+CRF&BiLSTM+CRF\cr
			layer & 1& 1&1&1\cr
			kernel& [3,4,5]&[3,4,5]&-&-\cr
			filter& 100&100&-&-\cr
			hidden size& 300&300&300&300\cr
			optimizer &Adam&Adam&SGD&SGD\cr
			dropout& 0.5&0.5&0.5&0.5\cr
			batch size & 50&50&768&768\cr
			epoch& 5&5&100&100\cr
			\bottomrule  
		\end{tabular}
		\caption{Hyperparameters for extrinsic datasets.}	
		\label{tab:hyper_parameter}
	\end{threeparttable} 
	%\end{minipage}%
\end{table}

\begin{figure}[tbp]
	\centering
	\includegraphics[width=0.5\textwidth]{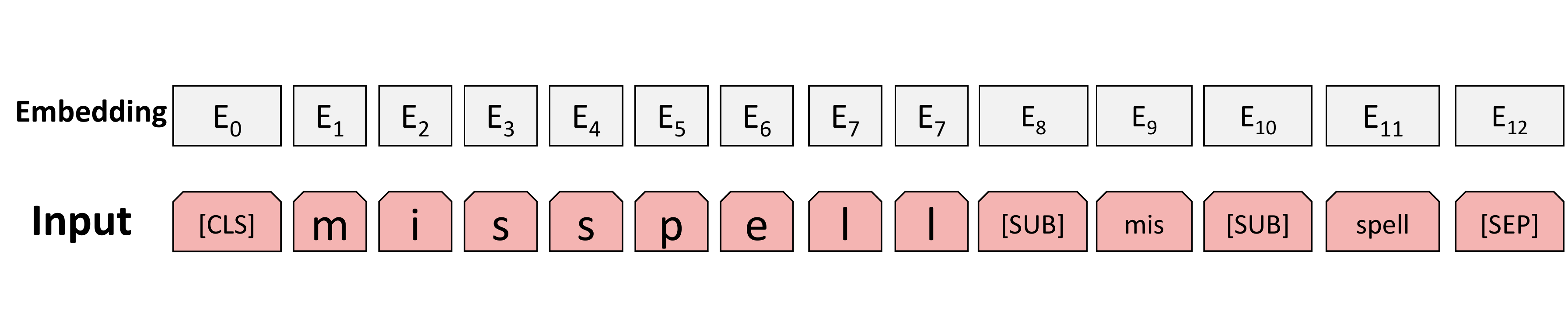}
	\caption{An illustration of our Mixed input for the word \texttt{misspell}.}
	\label{fig:mixed_input}
\end{figure}

\begin{figure*}[ptb]
	\centering
	\includegraphics[width=1.0\textwidth]{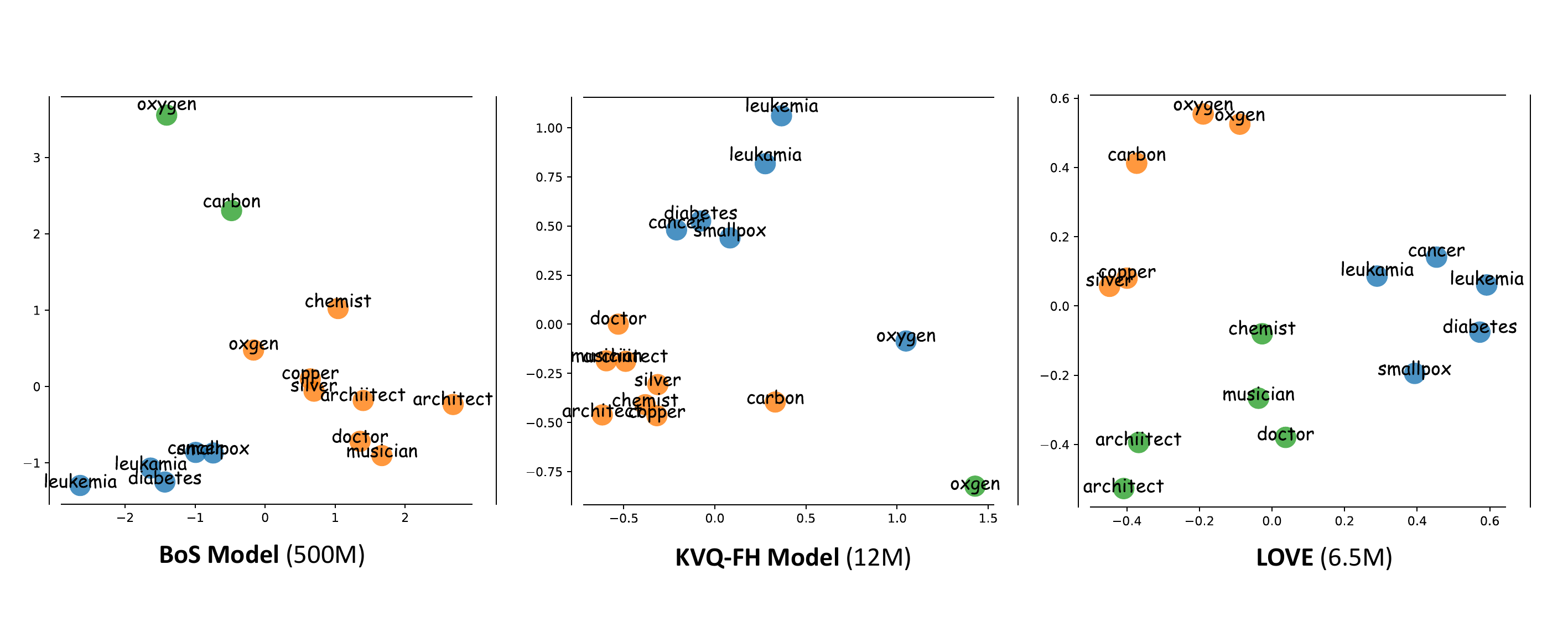}
	\caption{PCA visualizations of word vectors generated by LOVE, BoS, and KVQ-FH. Different colors mean different clusters, as predicted by K-means. There are three OOV words: \texttt{oxgen}, \texttt{archiitect} and \texttt{leukamia}.} 
	\label{fig:other_cluster}
\end{figure*}

\section{Details of Our Experiments} \label{sec:detail_experiment}
\subsection{Training of Mimick-like Models}
Our target pre-trained embeddings are those from FastText \texttt{fasttext-crawl-300d-2M}, because they provide a strong baseline. They sum up subword-level information to produce vectors for arbitrary words.
We also compare to MIMICK, BoS, and KVQ-FH, which do not train on contextual words.
We do not compare to Robust Backed-off Estimation~\cite{fukuda2020robust} and PBoS~\cite{jinman2020pbos}, because they need  larger and more complex models.
Robust Backed-off Estimation uses string matching to find the top-k similar words from the entire vocabulary when imputing.
Using the same target vectors, the number of parameter of BoS and PBoS are 163M and 316M, respectively.
We re-train MIMICK, BoS, and KVQ-FH as baselines according to the published settings.
In order to compare at the same parameter level, we use subwords for MIMICK instead of pure characters and adjust the hashing size $H=40K$ for KVQ-FH.

\subsection{Robustness Evaluations}
\begin{figure}[ptb]
	\centering
	\includegraphics[width=0.5\textwidth]{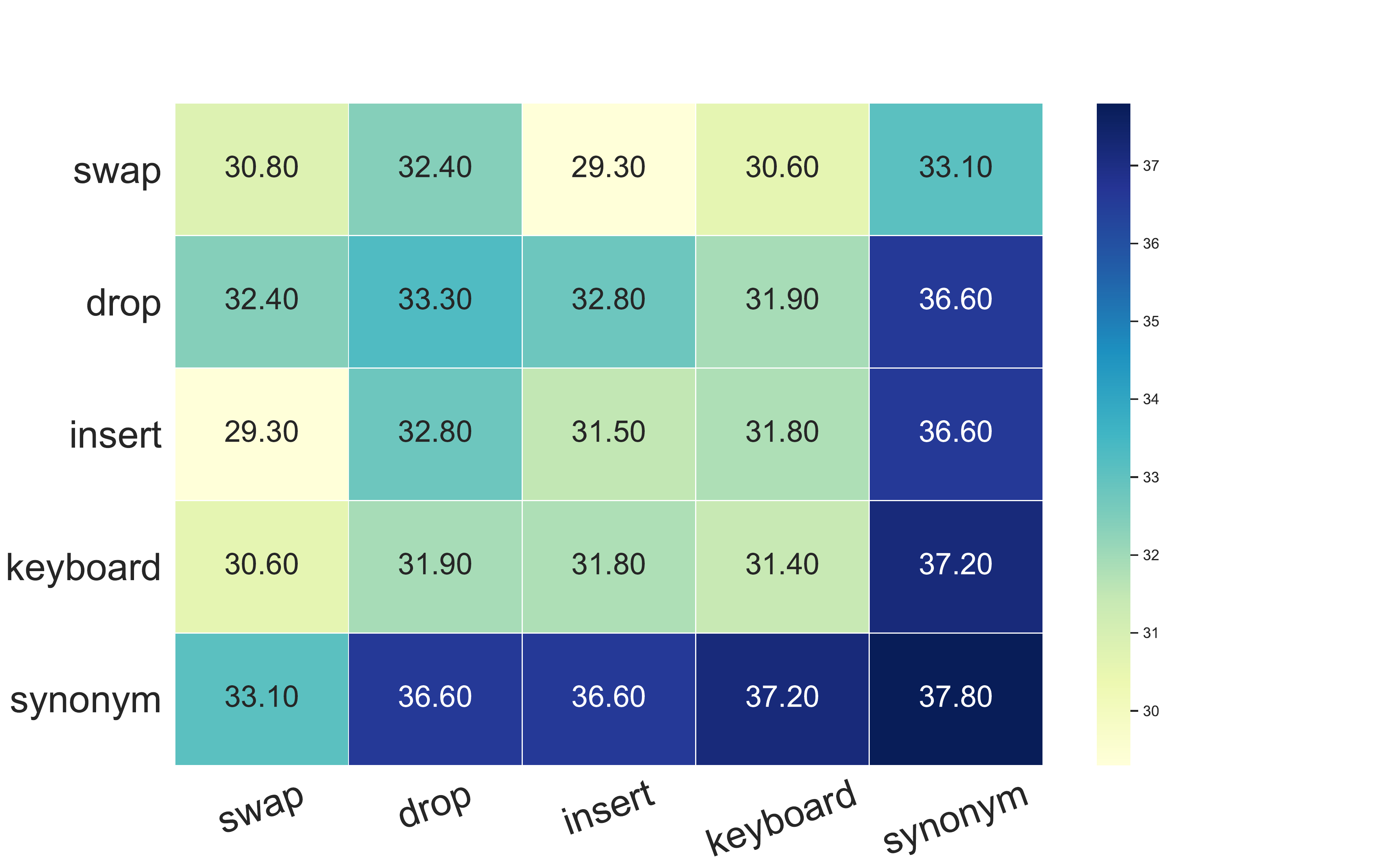}
	\caption{Performances of different augmentations on RareWord, measured as Spearman’s $\rho$. Diagonal entries correspond to individual augmentation and off-diagonal entries correspond to composite augmentation. }
	\label{fig:aug_on_rw}
\end{figure}

As for our model, we first lower-case and 
tokenize each word by using WordPiece~\cite{wu2016google} with a vocabulary of $30K$ subwords and preprocess them by stemming and removing subwords with numerals.
This yields a vocabulary of $21257$ words. 
Each subword is represented by corresponding vectors from our model and we adopt a modified attention model to encode the subword sequence.
Specifically, the layer number of this encoder is just $1$ for efficiency and the hidden dimension is $300$. 
In each block, the number of attention heads is $1$ and we use fixed sinusoidal position embeddings~\cite{vaswani2017attention} for positional information.
To train the contrastive learning framework, we use the open-source tool~\cite{ma2019nlpaug} to augment a word, and use the probabilities $\{0.07, 0.07, 0.07, 0.07, 0.36, 0.36\}$ for six augmentations:
swap, drop, insert, keyboard, synonym, no-operation. 
Hard negatives are generated by edit distance. For each target word, we store the top-100 similar words and insert $3$ of them into a mini-batch as hard negatives.
The loss function is a standard contrastive loss with temperature $\tau=0.07$. The optimizer is Adam and the learning rate is $0.002$. The dropout rate is $0.2$ and we train the model for $20$ epochs in total.

\subsection{Intrinsic and Extrinsic Evaluations}

We choose the setting discussed in Section~\ref{sec:approach} to train our model for 20 epochs, and evaluate each intrinsic task based on the vectors that the models produce.
As for the extrinsic tasks, we feed word vectors into each neural network and fix them during training.
We use CNNs for text classification~\cite{zhang2015sensitivity} and BiLSTM+CRF for NER~\cite{huang2015bidirectional}. 
We compare different embeddings on both intrinsic and extrinsic datasets by using generated vectors.
For the word cluster tasks, the produced vectors are clustered by K-Means and then measured by Purity.
The hyper-parameters of the extrinsic tasks are shown in Table~\ref{tab:hyper_parameter}.
For each dataset, our model is trained with five learning rates $\{ 5e-3, 3e-3, 1e-3, 8e-4, 5e-4 \}$. We select the best one on the development set to report its score on the test set.

To generate a corrupted dataset, we simulate post-OCR errors. %to add typos to each sentence.
We adopt the augmentation tool developed by~\citet{ma2019nlpaug} to corrupt $70\%$ of the original words. 
To check the robustness of BERT, we directly finetune a BERT-base model using Huggingface~\cite{wolf2020transformers}.
During finetuning, the batch size is $16$ and we train $5$ epochs.
We select the best model among five learning rates $\{ 9e-5, 7e-5, 5e-5, 3e-5, 1e-5\}$ on the development set
and report the score of the model on the test set.

\begin{figure}[ptb]
	\centering
	\includegraphics[width=0.5\textwidth]{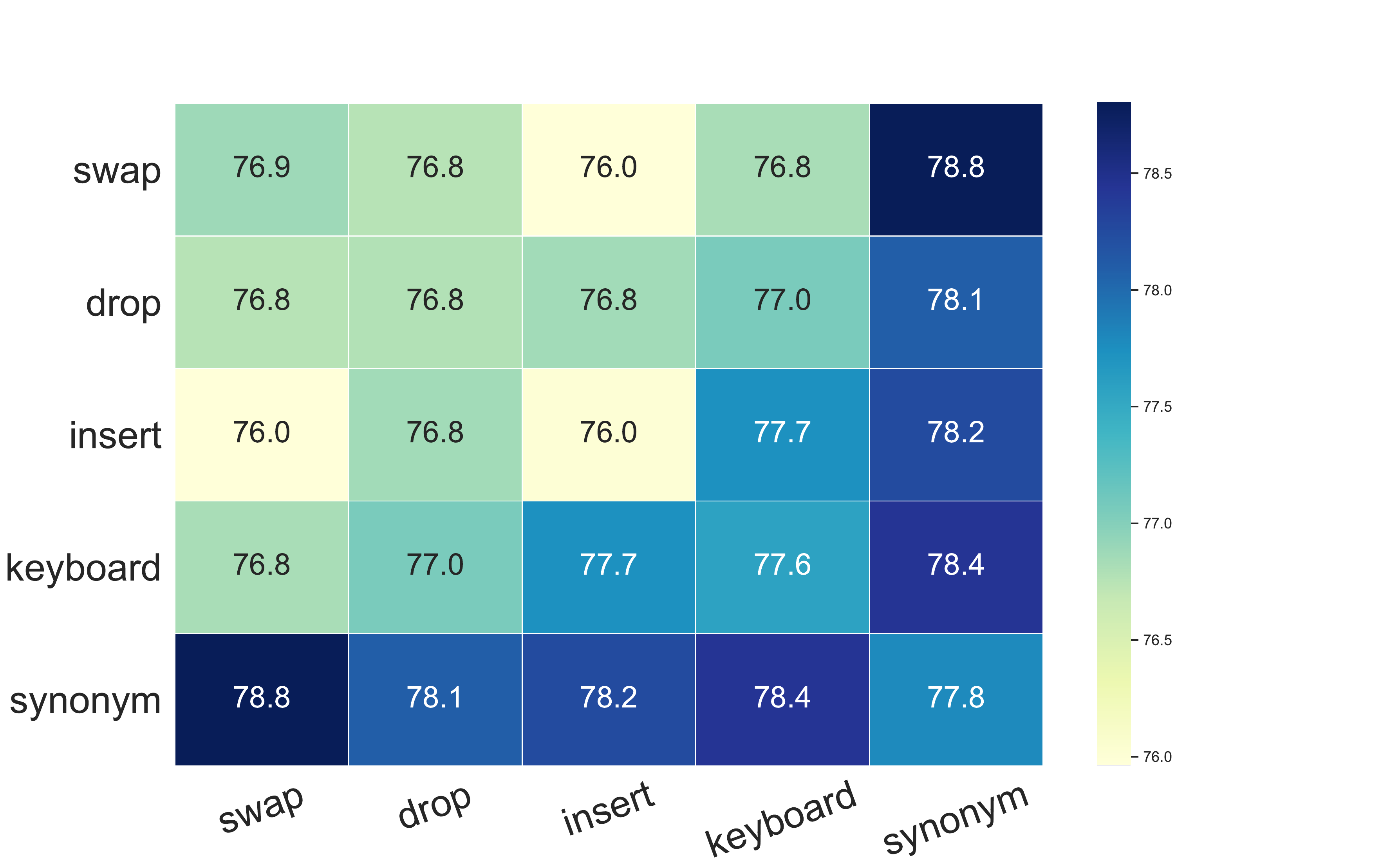}
	\caption{Performances of different augmentations on SST2, measured as accuracy. Diagonal entries correspond to individual augmentation and off-diagonal entries correspond to composite augmentation. }
	\label{fig:aug_on_sst2}
\end{figure}

\subsection{Datasets}\label{sec:detail_dataset}

\paragraph{Intrinsic Datasets.}
We use six word similarity datasets:
(1) RareWord~\cite{luong2013better}
(2) SimLex~\cite{hill2015simlex}
(3) MTurk~\cite{halawi2012large}
(4) MEN~\cite{bruni2014multimodal}
(5) WordSim~\cite{agirre2009study}, and 
(6) Simverb~\cite{agirre2009study}.
The task is scored by Spearman’s $\rho$, which computes the correlation between gold similarity and the similarity obtained from generated vectors.
For the word cluster task, we use
(1) AP~\cite{almuhareb2006attributes} and
(2) BLESS~\cite{baroni2011we}.
The generated word vectors are first clustered by K-means~\cite{macqueen1967some} and then scored by the cluster purity. 

\paragraph{Extrinsic Datasets.}

We use both sentence-level and token-level downstream datasets to evaluate the quality of word representations.
For the sentence level, we use SST2~\cite{socher2013recursive} and MR~\cite{pang2005seeing}, and the metric is accuracy.
For the token level, we use two NER datasets: general CoNLL-03~\cite{sang2003introduction} and biomedical BC2GM~\cite{smith2008overview}.
The metric is the entity-level F1 score.
As before, we select the best model among five different learning rates $\{ 5e-3, 3e-3, 1e-3, 8e-4, 5e-4\}$ on the development set and then report the model score on the test set.

\begin{figure}[tp]
	\centering
	\includegraphics[width=0.43\textwidth]{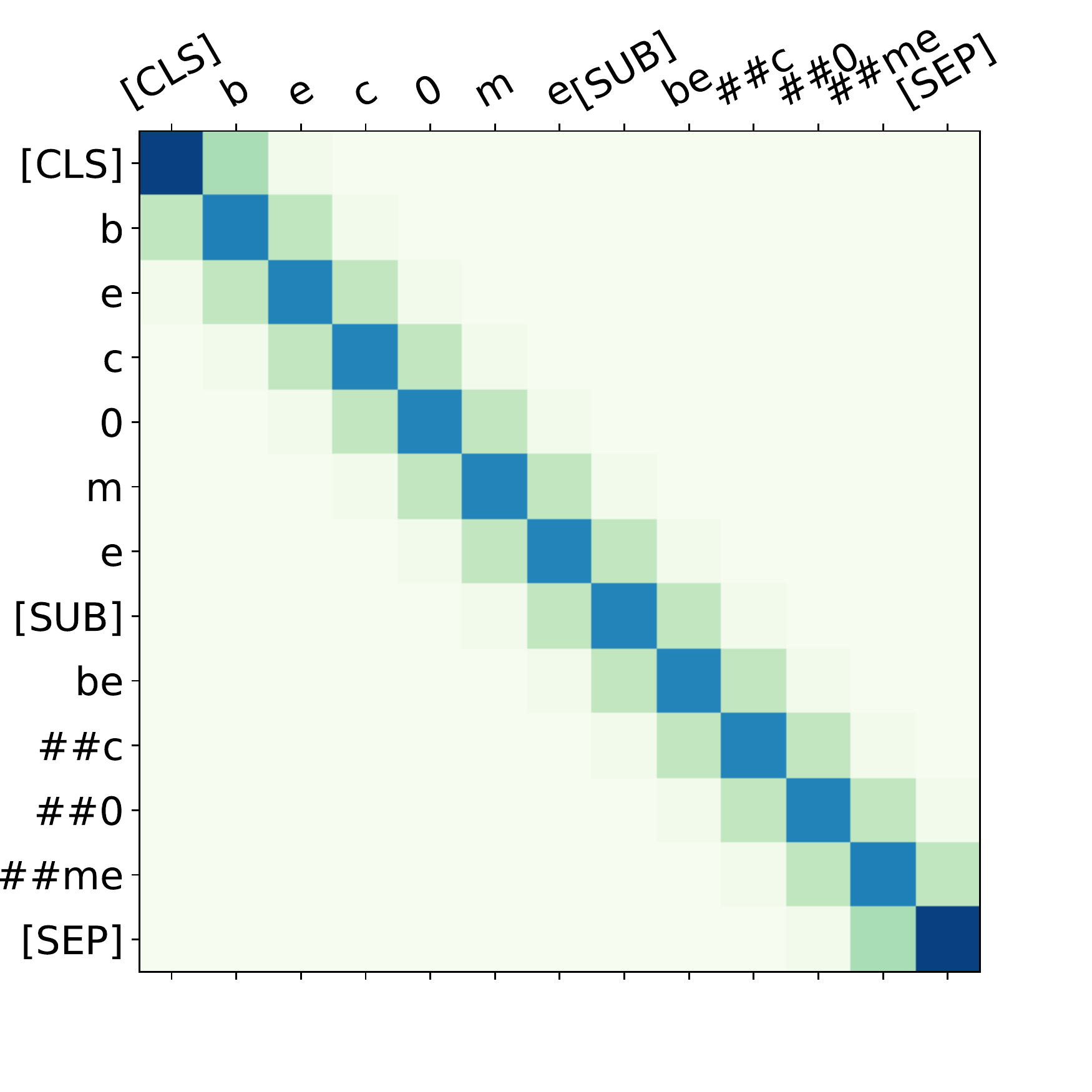}
	\caption{Visualization of positional weights for the post-OCR word \texttt{bec0me} (the correct one is \texttt{become}).} 
	\label{fig:visualization_pos}
\end{figure}

\begin{figure}[tp]
	\centering
	\includegraphics[width=0.45\textwidth]{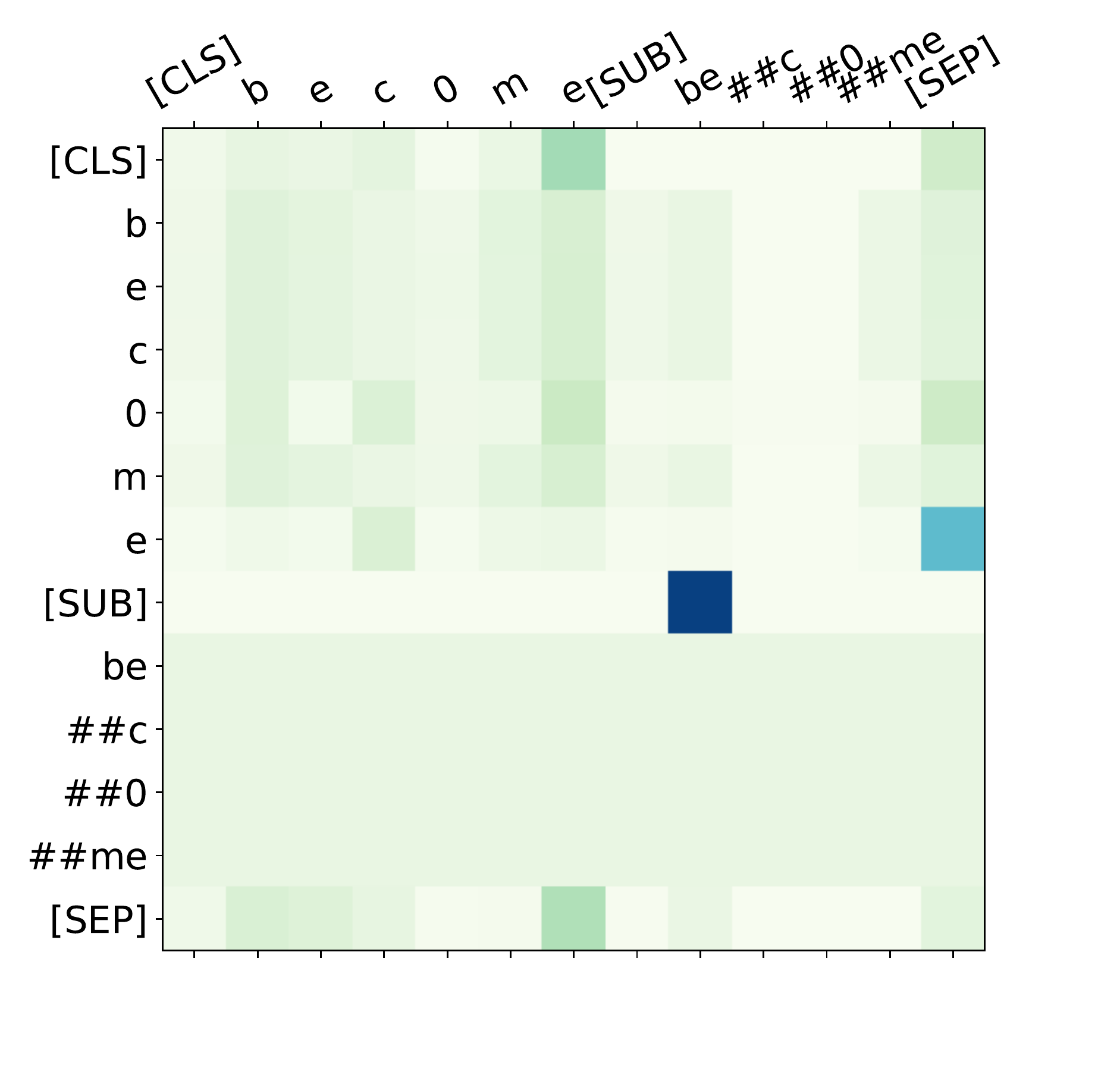}
	\caption{Visualization of self-attention weights for the post-OCR word \texttt{bec0me}.} 
	\label{fig:visualization_self}
\end{figure}

\section{Additional Analyses} \label{sec:analysis}
\subsection{Qualitative Analysis}
To better understand the clusterings produced by LOVE, we chose 15 words from the AP dataset~\cite{almuhareb2006attributes},
%. As shown in Figure~\ref{fig:other_cluster}, there are fifteen words that can be divided into 
covering three topics (Chemical Substance, Illness, and Occupation). We added 3 corrupted words, \texttt{oxgen}, \texttt{archiitect} and \texttt{leukamia}.
Figure~\ref{fig:other_cluster} shows how LOVE, BoS, and KVQ-FH cluster these words (using a PCA projection and K-means). All approaches space out the clusters to some degree. In particular, BoS and KVQ-FH have trouble separating professions and chemical substances. For the corrupted words, only LOVE is able to embed them close enough to their original form, so that they appear in the correct cluster. %that but both  cannot cluster them well and distinguish typo words well.
%For example, the word \texttt{oxygen} and \texttt{oxgen} should be close to each other and have the same cluster label, but both two models do not capture this feature well.

\subsection{Effect of Augmentation for Text Classification}

Figure~\ref{fig:aug_on_sst2} shows the performance of five augmentation strategies on the text classification task SST2.
We observe that synonym is the most effective methods. 
The first four methods have a weaker effect, but the keyboard replacement can bring a certain degree of improvement. The results on RareWord are similar (Figure~\ref{fig:aug_on_rw}).

\subsection{Effect of $\tau$ in Contrastive Loss}

As discussed in~\citet{chen2020simple}, a proper temperature can yield better representation in the visual area because $\tau$ is able to weigh the negatives by their relative hardness.
As shown in Table~\ref{tab:cl_tau}, we attempt different values of temperature and find that there is no consistent $\tau$ that makes a model work well both on intrinsic and extrinsic datasets.
Hence, we choose the best performer on average, i.e., $\tau = 0.07$.

\begin{table}[bt] 
	\centering 
	\tiny
	\begin{threeparttable} 
		\begin{tabular}{ccccc}  
			\toprule  	&\multicolumn{2}{c}{parameters}&RareWord&SST2\cr
			 &embedding&non-embedding&\cr
			\midrule
			$\ell_{cl}$ ($\tau$ = 0.03)&9M&200K&35.0&\textbf{81.6}\cr
			$\ell_{cl}$ ($\tau$ = 0.07) &9M&200K&39.8&81.3\cr
			$\ell_{cl}$ ($\tau$ = 0.12)&9M&200K&\textbf{39.9}&81.1\cr
			$\ell_{cl}$ ($\tau$ = 0.20)&9M&200K&37.6&81.5\cr
			$\ell_{cl}$ ($\tau$ = 0.50)&9M&200K&38.3&80.6\cr
			\bottomrule  
		\end{tabular}
		\caption{Performances of contrastive loss with various temperature $\tau$, measured as Spearman’s $\rho$ and accuracy respectively.}	
		\label{tab:cl_tau}
	\end{threeparttable} 
	%\end{minipage}%
\end{table}

\subsection{Visualization of Encoder}\label{sec:app_encoder}

As mentioned before, we combine two types of attention heads (self-attention and positional attention) to encode a subword sequence.
Here, we visualize the attention weights on each side and show how they work.
Figure~\ref{fig:visualization_pos} shows the position-dependent weights.
We use sinusoidal functions to generate positional embeddings, and the weights are the dot product between these embeddings.
We observe the positional weights tend to the left and right subwords in addition to themselves, which yields trigram representations.

Figure~\ref{fig:visualization_self} shows the self-attention weights which are computed from the trigram subwords of positional attention. 
Hence, each subword in this figure is a trigram representation instead of a single subword representation.
As we see, self-attention can capture global features regardless of distance.
We take the first token \texttt{[CLS]} as an example, and this self-attention assigns high weights for the token \texttt{e} and \texttt{[SEP]}, which constructs a representation like this: \texttt{[CLS]b} $+$ \texttt{me[SUB]} $+$ \texttt{\#\#me[SEP]}.
This segment tells us this word starts with \texttt{b} and ends with \texttt{me}.

\end{document}